\documentclass[acmsmall]{acmart}

\usepackage{caption}
\usepackage{subcaption}
\usepackage{hhline}
\usepackage{hyperref}
\usepackage{longtable}
\AtBeginDocument{%
  \providecommand\BibTeX{{%
    \normalfont B\kern-0.5em{\scshape i\kern-0.25em b}\kern-0.8em\TeX}}}

\setcopyright{acmcopyright}
\copyrightyear{2021}
\acmYear{2021}
\acmDOI{10.1145/1122445.1122456}

\acmJournal{THRI}
\acmVolume{37}
\acmNumber{4}
\acmArticle{111}
\acmMonth{8}

\begin{document}

\title{Using Design Metaphors to Understand User Expectations\\ of Socially Interactive Robot Embodiments}

\author{Nathaniel Dennler}
\email{dennler@usc.edu}
\author{Changxiao Ruan}
\email{changxir@usc.edu}
\author{Jessica Hadiwijoyo}
\email{hadiwijo@usc.edu}
\author{Brenna Chen}
\email{brennajc@usc.edu}
\author{Stefanos Nikolaidis}
\email{nikolaid@usc.edu}
\author{Maja Matari\'c}
\email{mataric@usc.edu}
\affiliation{%
  \institution{University of Southern California}
  \streetaddress{941 Bloom Walk, Los Angeles, CA 90089}
  \city{Los Angeles}
  \state{California}
  \country{USA}
  \postcode{90089}
}
\renewcommand{\shortauthors}{Dennler, et al.}

\begin{abstract}
 The physical design of a robot suggests expectations of that robot's functionality for human users and collaborators. When those expectations align with the true capabilities of the robot, interaction with the robot is enhanced. However, misalignment of those expectations can result in an unsatisfying interaction. This paper uses Mechanical Turk to evaluate user expectation through the use of design metaphors as applied to a wide range of robot embodiments. The first study (N=382) associates crowd-sourced design metaphors to different robot embodiments. The second study (N=803) assesses initial social expectations of robot embodiments. The final study (N=805) addresses the degree of abstraction of the design metaphors and the functional expectations projected on robot embodiments. Together, these results can guide robot designers toward aligning user expectations with true robot capabilities, facilitating positive human-robot interaction.
\end{abstract}

\begin{CCSXML}
<ccs2012>
<concept>
<concept_id>10003120.10003121.10003126</concept_id>
<concept_desc>Human-centered computing~HCI theory, concepts and models</concept_desc>
<concept_significance>500</concept_significance>
</concept>
<concept>
<concept_id>10003120.10003121.10011748</concept_id>
<concept_desc>Human-centered computing~Empirical studies in HCI</concept_desc>
<concept_significance>500</concept_significance>
</concept>
</ccs2012>
\end{CCSXML}

\ccsdesc[500]{Human-centered computing~HCI theory, concepts and models}
\ccsdesc[500]{Human-centered computing~Empirical studies in HCI}
\ccsdesc{Computer systems organization~Robotics}
\keywords{Socially Interactive Robots, Robot Morphology, Social Perceptions}

\maketitle

\begin{figure}[h]
  \centering
  \includegraphics[width=\linewidth]{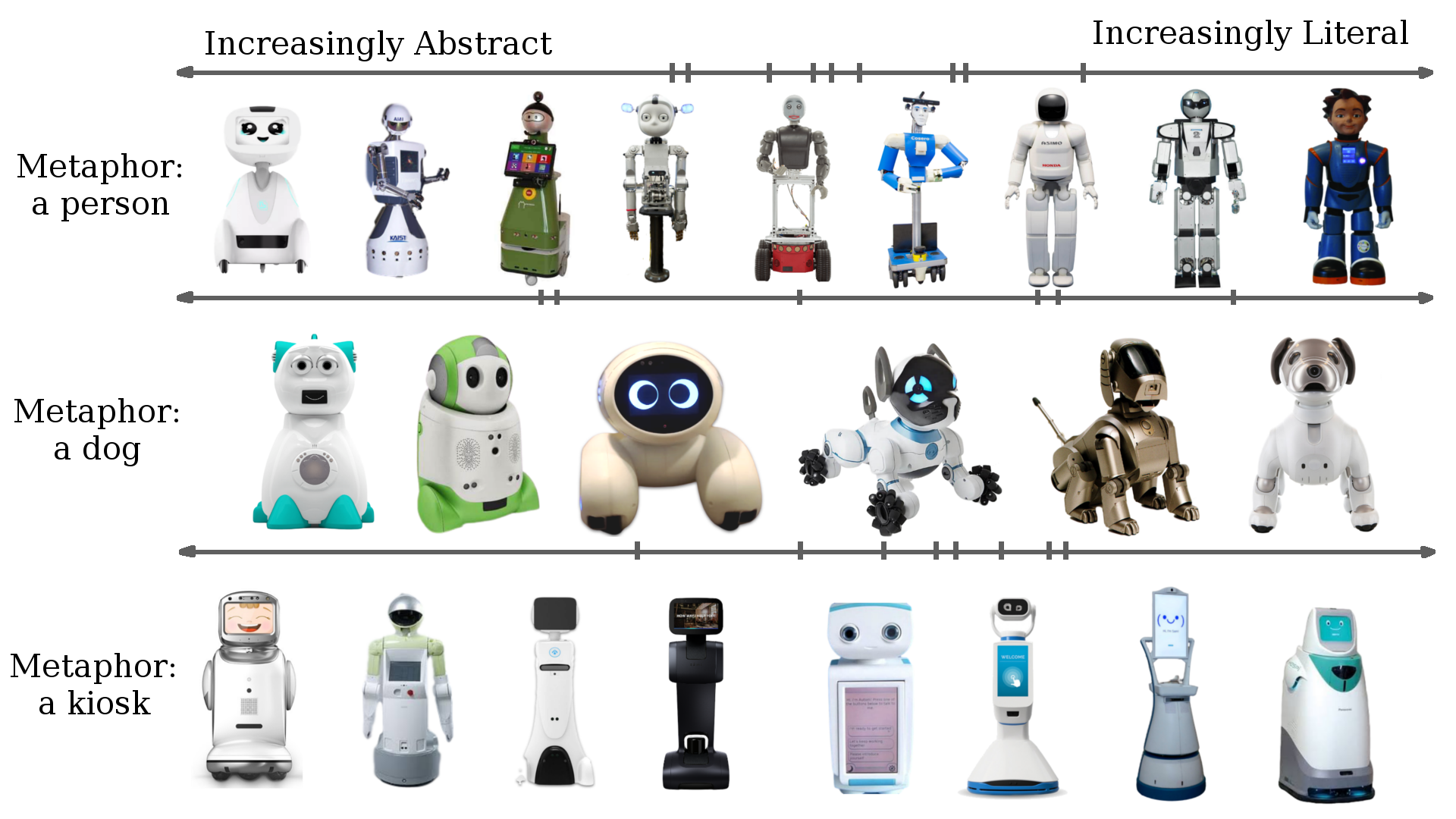}
  \caption{Examples of robots' physical designs measured by abstraction level along three different design metaphors.}
  \label{illustration}
  \Description{The image shows three axes, each labelled from increasingly abstract on the left to increasingly literal on the right. The first axis is labeled "Metaphor: a person" and shows nine robots ranging from robots that use wheels to move on the left to robots that have arms and legs on the right. The second axis is labeled "Metaphor: a dog" and shows six robots ranging from robots that sit upright on a table on the left to robots that have four legs and dog-like ears on the right. The third axis is labeled "Metaphor: a kiosk" and shows robots that range from robots that have human-like qualities but are box shaped on the left to robots that look like ATM machines (kiosks) on the right.}
\end{figure}

\section{Introduction}

Broad acceptance of socially interactive robots by users is a barrier to increasing their prevalence in daily life. One of the key factors that drives acceptance of new technologies is user expectations \cite{cha2015perceived,davis1989perceived}. In addition to perceived usefulness, extensive past research has shown that users consider some computer and robot systems as social actors \cite{nass1994computers,carpinella2017robotic,shadbolt2013towards,mathur2016navigating} and thus assign social judgements to these systems in addition to judgements of purely functional affordances. The formation of these social and functional expectations is crucial to understanding how to design effective human-robot interactions (HRI) for various real-world contexts.
 
Physical embodiment is fundamental to robotics and HRI. Embodiment has been studied in several fields, including social psychology \cite{meier2012embodiment}, cognitive sciences \cite{varela2016embodied}, and philosophy \cite{anderson2003embodied}. Embodiment broadly refers to the agents' structural coupling to their physical environment \cite{ziemke2001disentangling}. In this work, {\it embodiment} is the robot's connection with the physical world, through sensors and actuators as the means of acting on the world, similar to the definition used by \citet{deng2019embodiment}. The physical robot embodiment conveys a sense of robot identity and allows the robot to become situated in social contexts. Socially interactive robots \cite{fong2003survey} are designed to work and interact with people in social contexts, and adapt to the social preferences of different users.

Adapting technology to the user has multiple benefits, including increasing trust \cite{nikolaidis2017human}, increasing intention to use \cite{tay2014stereotypes}, and improving behavioral quantities of interest such as time spent in the interaction \cite{goetz2003matching, tapus2008user, cruz2016teaching}. Some aspects of a robot's perceived identity are readily changed over time with few noticeable inconsistencies, such as expressing personality through gaze patterns \cite{andrist2015look}, prosodic features \cite{fischer2019speech}, and gesture \cite{rifinski2020human}. Other aspects of a robot's perceived identity, however, can be disruptive if changed, such as voice pitch \cite{tapus2008user}. Physical design is difficult or impossible to change over the course of an interaction without significant expense and disruption. Thus, robot design is a complex process of optimizing over both fixed physical design and changing social presence \cite{deng2018formalizing}. An effective design team must jointly explore the coupled spaces of embodiment and personality. If those design spaces are large, finding designs to satisfy goals becomes costly while if they are small, potentially effective solutions may be excluded.

% This paper aims to add structure to the space of socially interactive robot physical morphologies by exploring the social and functional cognitive biases that users place on different robot embodiments. Section \ref{relatedWork} discusses related work. Section \ref{datasetFormulation} describe the methodology behind the selection of robots in the dataset and attributing design metaphors to those robots. Sections \ref{socialPerception} and \ref{functionalPerception} outline the Amazon Mechanical Turk user studies addressing the social and functional perceptions of the presented robots. Section \ref{exploration} proposes ways to visualize the space of embodiments and analyzes the results, Section \ref{discussion} offers insights into the design of socially interactive robot embodiments, and Section \ref{conclusion} concludes the paper.

This work contributes an open-source database of 165 robot embodiments with associated descriptors and results of three crowd-sourced studies that provide insights toward the effect of robot design on user expectations of robot capabilities. Section~\ref{methodology} describes the crowd-sourced data collection process through Mechanical Turk. The first study (N=382) associates crowd-sourced design metaphors to different robot embodiments as illustrated in Figure \ref{illustration}. The second study (N=803) assesses initial social expectations of robot embodiments. The final study (N=805) addresses the degree of abstraction of the design metaphors and the functional expectations projected on robot embodiments. Section~\ref{overallFindings} examines trends in the design of socially interactive robots as a whole. To evaluate the usefulness of the database, we show how viewing robots through design metaphors can contextualize and extend prior work in HRI, focusing on social expectations in Section~\ref{socialExpectations} and functional expectations in Section~\ref{functionalExpectations}. We discuss the implications of using metaphors as a tool for both designers of interactions and designers of robots in Section~\ref{discussion}. To support replicable research, the collected dataset and interactive explorations of the dataset are available at~\href{interaction-lab.github.io/robot-metaphors/}{ \texttt{interaction-lab.github.io/robot-metaphors/}}.
\section{Background and Related Work}
This section provides background about research through design and about design metaphors for setting user expectations. Our work extends those methods toward studying the formation of user expectation about a robot's embodiment. We additionally present a review of work that aims to achieve similar goals regrading robot design and identify how our contributions can contextualize findings from past works.
% \subsection{The Technology Acceptance Model}
% Mental Models
% TAM

\subsection{Research Through Design}
% Talk about the pragma of doing research through design, citing the papers from 2020 HRI. Specifically talk about intermediate level design knowledge.

%While randomized controlled trials (RCT) are the gold standard for causal hypothesis testing, the act of crafting such a highly-controlled studies requires that the resulting finding is couched within the prespecified, controlled context. Unfortunately, real-world scenarios present numerous uncontrollable variables, so some RCT findings may not be representative of subsequent real-world deployments. Interactions with people involve confrontations with innumerable \textit{wicked problems} \cite{rittel1973dilemmas} characterized by their inability to be perfectly described {\it a priori}.  reductionist approach.

The paradigm of \textit{research through design} (RtD) \cite{zimmerman2007research} provides a method for tackling the complexity of designing for real-world interactions. The paradigm highlights the importance of the exploratory implementation of systems to solve real-world scenarios. These specific implementations are called \textit{design artifacts} and their production creates knowledge of different design patterns, design processes, and other forms of design knowledge that may be useful in diverse settings. 

The RtD paradigm has been explored in robot design. Examples include prototyping practices in virtual, physical, and video-based contexts \cite{ju2021fake}. Another avenue of RtD is through the documentation of participatory design approaches with specific end-user populations, such as designing creative robots with children \cite{alves2021children} and designing augmented communication robots with users that have cerebral palsy \cite{valencia2021co}. To properly evaluate the produced design artifacts, an inspection of several disparate artifacts is required, in order to identify trends and patterns across many different contexts. This meta-knowledge across design artifacts is termed intermediate-level design knowledge \cite{hook2012strong}. Due to their structural coupling to the real world, robots in particular have unique characteristics and limitations that allow for the formulation of different forms of design knowledge \cite{lupetti2021designerly}. Recently, emphasis has been placed on using design techniques from data visualization to properly interpret robot intentions as they interact with humans \cite{szafir2021connecting}. The use of the RtD paradigm provides an expanded understanding of what users expect when interacting with various robots in different contexts. The dataset and visualization techniques used in this work serve as a form of intermediate-level design representation of the space of embodiments, similar to other commonly used representations, such as annotated design portfolios \cite{gaver2012annotated}.

\subsection{Understanding Design Through Metaphors}
{\it Design metaphors} concisely describe complex ideas by associating unfamiliar objects with other objects that have similar characteristics the user has already experienced. Design metaphors are extensively studied in human-computer interaction (HCI) as a way to help users develop mental models of the system they are interacting with \cite{voida2008re, jung2017metaphors, khadpe2020conceptual, kim2020conceptual}. For example, HCI research shows that metaphors shape user perceptions of fundamentally identical chat-bot systems  with different metaphors, resulting in varying levels of perceived warmth and competence affecting both the users' pre-interaction intention to use the system and their intention to adopt the system post-interaction~\cite{khadpe2020conceptual}.

The notion of design metaphors has also been recently applied to formalizing general design processes for socially interactive robots \cite{deng2018formalizing}. \citet{deng2019embodiment} provides a comprehensive review of HRI studies through the lens of design metaphors of the robot embodiments used and provides a design metaphor-based analysis of the relationships between different studies and their outcomes \cite{deng2019embodiment}. We apply this framework to explore how design metaphors shape the formulation of social and functional expectations of robots.
% This work aims to connect the RtD paradigm and the design metaphor methodology to form intermediate-level design knowledge of socially interactive robots to guide design decisions made by HRI researchers through the lens of design metaphors.

% \section{Related Work}
% We present a review of work that aims to achieve similar goals regrading robot design and identify how our contributions can explain and contextualize findings from past works.

\subsection{The Social Perception of Robot Embodiment:}
Past work in perception of robot embodiments is extensive, but is typically limited to few embodiments in a few contexts \cite{walters2007exploring,ventre2019embodiment}. One study found that a human-like embodiment is perceived as more competent than a speaker embodiment even when both embodiments fail in a cooking instruction task \cite{kontogiorgos2020embodiment}. Another study reported on relationships between the effects of an anthropomorphic, a zoomorphic, and a functional embodiment across four different tasks on social measures including trust, likeability, and engagement \cite{li2010cross}.

The effect of robot embodiments on their perceived gender has been studied in depth \cite{eyssel2012s}. One study found that certain physical robot characteristics, such as a particular waist-to-hip ratio, have a significant impact on the perceived gender \cite{trovato2018she}. Other studies showed that the combination of robot gender and occupation affect the perceived competence of the robot and the user's trust in the robot to accomplish occupational tasks \cite{bryant2020should}.  

The uncanny valley and robot likeability more generally have also been extensively explored  \cite{mathur2016navigating,strait2017public,woods2004design}. One study aimed to link aversion related to the uncanny valley to the ambiguity and atypicality of a robot's design \cite{strait2017understanding}. Robots that have more ambiguous and atypical embodiments were shown to be perceived as more eerie, which can have implications on design metaphors in general. Another study showed similar effects with zoomorphic types of embodiments and offered empirical evidence that something similar to the uncanny valley effect also happens with animals \cite{loffler2020uncanny}. While many studies group social expectations of large groups of robots along a single axis (e.g., human-likeness or animal-likeness), our work aims to explain social expectation of many distinct robots through the use of design metaphors.

\subsection{Social Perception of Robot Faces}
The design of robot faces specifically has been an area of research focus, as the face is often one of the most salient aspects of a socially interactive robot. McCloud's triangular design space for robot face design is used to contextualize design decisions \cite{blow2006art}. A comprehensive survey by \citet{kalegina2018characterizing} of screen-rendered robot faces reports on commonly-used design patterns, and the features that correspond with relevant social characteristics of those faces. Our work aims to show common patterns in the design space of robot embodiments and to clarify how they impact social and functional expectations of robots.
% \section{Dataset Formulation}
% Incorporated into the studies section since it was only one subsection
\section{Collecting Crowd-Sourced Data on User Perceptions} \label{methodology}

%We collected several measures of user expectation of different robots. 
The dataset we collected aims to describe robots holistically in terms of design metaphors as well as users' expectation of the robots' social and functional capabilities. To minimize participant fatigue due to the large number of measures we collected, we split the data collection over three separate studies. All studies were conducted on Amazon Mechanical Turk (MTurk). 

\subsection{Data Quality}\label{quality}

To maximize the quality of collected responses, we used the following inclusion criteria for all studies: user approval ratings of $\geq 99\%$, $\geq 1000$ completed tasks, and normal or corrected to normal vision.  Furthermore, we limited the participants to residents of the United States in order to control for cultural factors of design metaphors. The full set of survey questions is presented in Appendices~\ref{DesignSurveyQuestions}-\ref{FunctionalSurveyQuestions}.

In addition to the inclusion criteria, several measures were taken throughout the survey to protect against non-human MTurk participants. We employed a "honeypot" question that was invisible on the survey, but visible through the HTML files, thereby identifying electronic participants when answered. Participants who responds to this question were excluded from the study. Additionally, we employed random "attention checks" that instructed users to answer a question by selecting a specific option to continue the survey, requiring careful reading of the question. For quantitative questions, if the user responded with all neutral responses, their survey ended early and the data were discarded. For qualitative questions, if the user responded with identical strings to ones that they had used previously (indicating copy-pasting), their survey ended early and the data were discarded.  This study was approved by the IRB (UP-18-00510); per IRB regulations and ethical considerations, participants were paid for the portion of the survey they completed.

\subsection{Data Collection Methodology}

\begin{figure}[ht]
  \centering
  \hspace*{\fill}%
  \begin{subfigure}[b]{0.45\textwidth}
         \centering
         \includegraphics[width=\textwidth]{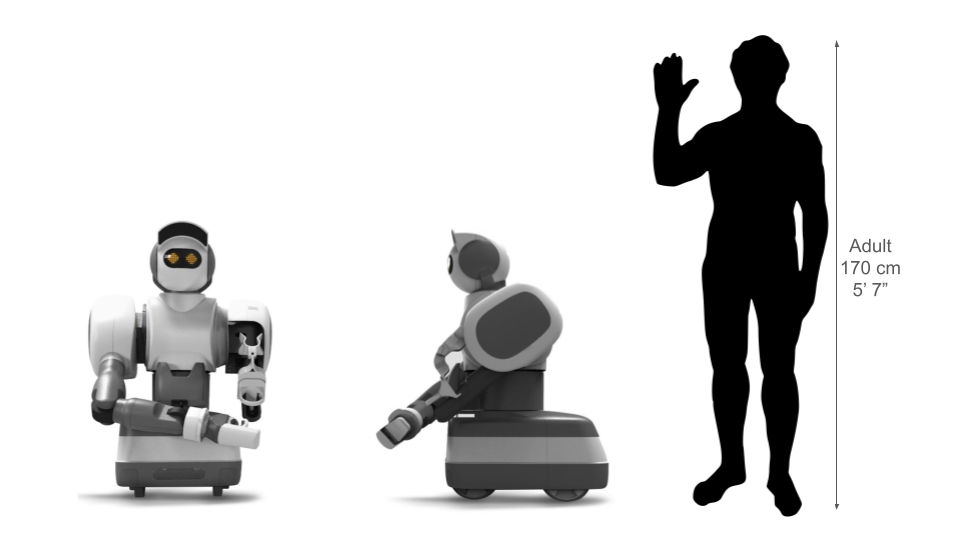}
         \caption{Stimulus for Aeolus.}
  \end{subfigure}
  \begin{subfigure}[b]{0.45\textwidth}
         \centering
         \includegraphics[width=\textwidth]{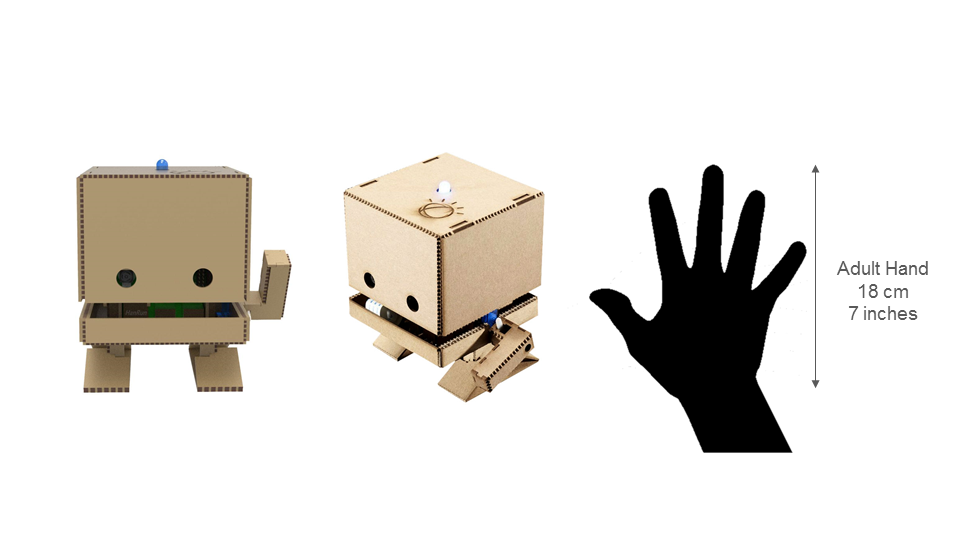}
         \caption{Stimulus for TJBot.}
     \end{subfigure}
  
  \caption{Example stimuli from the database.}
  \Description{}
   \label{fig:aeolusslide}
\end{figure}

Due to the immense cardinality of the design space of robots in widely varied contexts (i.e., drones, autonomous vehicles, industrial robots, etc.), we limited the scope of our dataset to robots that fit the definition of a {\it socially interactive robot} as proposed by \citet{fong2003survey}, with the exception of the requirement for high-level dialogue, in order to include more non-humanoid embodiments. Using those guidelines, we assembled a collection of 165 robots from the IEEE "ROBOTS: Your guide to the world of robots" site \cite{spectrum_2018} and Google searches of "Social Robot", "Socially Interactive Robot", "Socially Assistive Robot", "Robot Pet", and "Social Robot Animal". Google searching was performed under several user profiles as well as in incognito modes to mitigate the effects of prior search histories and stored user information. The data collection took place in June of 2020.

Each robot was presented to the study participants as two high-quality images, one of a front view and one of a side view, to convey the 3D structure of the robots' design. The sense of scale was provided by showing a common reference photo: a gender-neutral silhouette for robots at/over 80 centimeters in height or a silhouette of an averaged-sized human hand for robots under 80 centimeters in height. The image backgrounds were solid white, to control for contextual factors, cues, and influence. In addition, any objects that a robot was holding in the original image were edited out; we prioritized the use of images of robots in neutral poses with neutral facial expressions. All images were created with identical aspect ratios and similar sizes for each view of each robot, with minimal necessary variations due to the size and structure of different robot embodiments. Two example stimuli from our database are shown in Figure \ref{fig:aeolusslide}.

\subsection{Study 1: Crowd-Sourcing Design Metaphors}
The goal of the first study was to crowd-source the design metaphors to be used as the basis for subsequent analyses. The interface for the study is provided in Appendix \ref{DesignSurveyQuestions}.
\newpage
\subsubsection{Study Design}\hfill

A total of 382 participants (full demographic information shown in Appendix~\ref{demographicsinfo}) took part in the study and were paid \$1.00 per robot for which they provided 2-5 design metaphors. Participants viewed up to five robots that were presented in a randomized and counter-balanced manner. Each valid response took a median value of 3 minutes, and the whole survey took around 15 minutes.

%We asked the participants for specific concepts that can be attributed to each robot. We additionally asked respondents to provide a description of the robot to understand what features of the robot were used to describe the robot.  The details of the measures are described next.

\subsubsection{Qualitative Measures}\hfill

\noindent 
\textit{Description of Robot:} We provided an open-form response box with the prompt to describe the robot to a friend using two to three sentences.

\noindent 
\textit{Related Design Metaphors:} We provided an open-form response box to input at least two and up to five specific persons, animals, plants, characters, or objects that the robot looks like.

\noindent 
\textit{Reasoning for Related Design Metaphors:} We provided an open-form response box to describe why the above design metaphors were chosen.  This box was immediately to the right of the previous response box. 

\subsubsection{Overview of Collected Data}\hfill

We collected 1716 responses from the participants for the 165 robots we collected. Answers that did not provide specific persons, animals, plants, characters, or objects (i.e., 'good', 'nice', or paragraphs of copy-pasted text) were excluded from analysis. The removal of these responses did not result in a difference from the original distribution of responses, as evidenced by a chi-square test $\chi^2$ (1, N = 165) = 12.62, $p > .999$. This indicates that the excluded answers followed a uniform distribution and that the assignment of robots did not cause the participants to provide non-specific answers.

\subsection{Study 2: Social Perception}

The goal of the second study was to measure the social attributes of the robots that form the user's expectation of how a robot should behave. The interface for the study is provided in Appendix \ref{SocialSurveyQuestions}.

\subsubsection{Study Design}\hfill

The study was first pilot-tested with 10 na\"ive users to confirm the reliability of the constructed scale and get a sense of the time to complete it in order to inform pricing, following recommendations from \cite{rueben2020introduction}. These users were excluded from the full study. 

A total of 803 participants (full demographic information shown in Appendix~\ref{demographicsinfo}) took part in the study. The study followed a mixed design where each participant provided ratings for up to five robots. Assignment of robots was randomized and counter-balanced. Each rating was paid \$0.20, and took a median of 1.5 minutes per robot for an expected maximum length of 7.5 minutes for the whole survey. 

\subsubsection{Quantitative Measures}\label{study2measures}\hfill

\noindent 
\textit{RoSAS Scale:} We used a modified version of the validated RoSAS scale \cite{carpinella2017robotic} to assess the constructs originally defined in RoSAS, which we deemed to be reliable from the pilot study. All items followed the prompt "Indicate how closely the following words are associated with the robot" and were rated on a 7-point Likert scale of "strongly disagree" to "strongly agree". The scale measures the following constructs:

\begin{itemize}
    \item {\it Warmth} is related to the perception that another agent may want to help or harm us.
    \item {\it Competence} is related to the perception that another agent has the ability to help or harm us.
    \item {\it Discomfort} is related to the awkwardness of a robot.
\end{itemize}

\noindent
\textit{Robot Gender Expression: }
While gender is a complex social phenomenon, we measured \textit{perceived gender expression} as proposed by the Bem Sex-Role Inventory Scale \cite{bem1981bem}, using two axes-- masculinity and femininity--as 7-point Likert scales. This approach allowed for perceptions of androgyny and agender in terms of the two axes.

\noindent
\textit{Social Role:}
The social role is a measure of the interaction dynamics between the person and robot in an interaction \cite{rae2013influence,deng2019embodiment}. The scale uses a 9-point differential scale, where 1 labels the robot as "a subordinate", 5 labels the robot as "a peer", and 9 labels the robot as "a superior".

\noindent
\textit{Identity Closeness:}
Identity closeness measures the degree of in-group identification of the person with the robot \cite{tajfel1974social}. The scale uses a 9-point differential scale where 1 corresponds to the rater viewing the robot as "not at all like me", and 9 corresponds to the rater identifying the robot as "exactly like me". This scale has been shown to achieve high validity and reliability in related contexts \cite{reysen2013further}.

\noindent
\textit{Likeability:}
Likeability measures the general attitude toward a robot, and has been used in other robot assessment studies \cite{kalegina2018characterizing, mathur2016navigating}. It is assessed using a 9-point differential scale, where 1 indicates the rater "strongly dislikes" the robot, and 9 indicates that the rater "strongly likes" the robot, from the Godspeed Scale \cite{bartneck2009measurement}.

\subsubsection{Qualitative Measures}\hfill

\noindent 
\textit{Reasoning for Likeability Rating:}
In addition to the likeability rating, we collected an optional open-ended response about the reasons for liking or disliking the robot.

\subsubsection{Overview of Collected Data}\hfill

We collected 3481 ratings from the participants for the 165 robots in the dataset. Participants who failed random attention checks ended the survey early.   Entries that provided nonsensical answers to the qualitative questions or behaved randomly on the questionnaire were excluded. A total of 3155 responses were ultimately included in the analysis. A chi-square test showed that the exclusion of these 326 responses did not significantly affect the uniform distribution of assignment, $\chi^2$ (1, N = 165) = 72.83, $p > .999$, indicating that it is unlikely that certain robots are more associated with excluded answers than other robots. The modified version of the RoSAS scale showed high reliability with Cronbach's alphas of $\alpha=0.84$ for Warmth, $\alpha=0.87$ for Competence, and $\alpha=0.81$ for Discomfort. The values of the four questions that measured each construct are averaged for analysis.

\subsection{Study 3: Functional Perception}

The goal of the third study was to measure the expected functional affordances of the robots. The interface for the study is provided in Appendix \ref{FunctionalSurveyQuestions}.

%we used a modified version of the EmCorp-Scale \cite{hoffmann2018peculiarities}; a scale that has been validated in online survey contexts. We are interested in the constructs of Shared Perception and Interpretation, Tactile Interaction and Mobility, and Nonverbal Expressiveness. We do not consider the Corporeality construct as these robots are all images and corporeality represents how co-present the robot is in the room with an observer. We additionally evaluate the abstraction level of the metaphors collected from the first study, and collect open-ended descriptions of tasks that each robot might be good for.

\subsubsection{Study Design} \hfill

A total of 805 participants (full demographic information shown in Appendix~\ref{demographicsinfo}) took part in the study. The study followed a mixed design where each participant provided ratings for up to five robots. Each rating that a participant gave was paid \$0.50. Each rating took approximately 2 minutes to provide, and the whole survey had an expected length of 10 minutes. The robots each participant saw were randomized and counter-balanced to mitigate ordering effects. 

\subsubsection{Quantitative Measures}\hfill

\noindent 
{\it EmCorp Measures:}
We used a modified version of the 7-point Likert EmCorp-Scale \cite{hoffmann2018peculiarities} that has been validated in online survey contexts. We are interested in the constructs of Shared Perception and Interpretation, Tactile Interaction and Mobility, and Nonverbal Expressiveness. We did not consider the Corporeality construct as the robots used in the study are all images and corporeality represents how co-present the robot is in the room with an observer. All items were rated on a scale from "strongly disagree" to "strongly agree". The scale measured the following constructs.

\begin{itemize}
    \item{\it Shared Perception and Interpretation} is a measure of the perceived perceptual capabilities (such as vision and hearing) of the robot. 
    \item {\it Tactile Interaction and Mobility} is a measure of the perceived ability of the robot to move around and manipulate objects in space.
    \item {\it Non-verbal Expressiveness} is a measure of the robots ability to use natural cues such as gestures and facial expressions.
\end{itemize}

\noindent
{\it Design Ambiguity and Design Atypicality Measures:}
Design ambiguity and atypicality have been linked to aversion toward different robot designs in previous works \cite{strait2017understanding}. In this work, we define {\it ambiguity} as the difficulty of placing a robot in a single category, and {\it atypicality} as a robot having embodiment features not usually associated with the category they represent. We quantify these measures with differential scales valued from 1 to 9. 

\noindent
{\it Metaphor Abstraction Measures:}
The abstraction level of a metaphor provides a way to quantify how abstractly or literally the robot embodiment follows the metaphor. We quantified these values as a 9-point differential scale where 1 represented "highly abstract" interpretations of the design metaphor, and 9 represented "highly literal" interpretations of the design metaphor.

\subsubsection{Qualitative Measures}\hfill

\noindent
{\it Task Descriptions:}
We required participants to report two to five kinds of tasks each robot would be appropriate for, using open-ended responses.

\subsubsection{Overview of Collected Data}\hfill

We collected 3435 ratings for the 165 robots in the dataset. Participants who failed random attention checks ended the survey early. Some responses were excluded for exhibiting disengaged or automated behaviors, as outlined in Section \ref{quality}. A total of 3092 responses were ultimately included in the analysis. A chi-square test confirmed that the exclusion of the 343 responses did not significantly affect the uniform distribution of assignment, $\chi^2$ (1, N = 165) = 25.39, $p > .999$, indicating that it is unlikely that certain robots are more associated with excluded answers than others. The modified version of the EmCorp-Scale showed high reliability with Cronbach's alphas of $\alpha=0.91$ for Shared Perception and Interpretation, $\alpha=0.84$ for Tactile Interaction and Mobility, and $\alpha=0.87$ for Nonverbal Expressiveness. The values of the four questions that measured each construct were averaged.
\section{Overall Findings in the Design Space} \label{overallFindings}

\subsection{Describing Embodiments Through Features} \label{datasetOverview}
Similar to \citet{kalegina2018characterizing}, we described the robot embodiments with a series of manually labeled features derived from observed design patterns of the robots in the dataset in conjunction with qualitative analyses of descriptions participants gave when describing the robot. In total, we created 43 binary or categorical variables related to present/absent features, 4 ordinal variables related to feature counts, and 5 continuous variables. The full set of coded features and their descriptions are shown in Appendix~\ref{RobotDescriptions}. Because these features were developed specifically for the robots in this database, they are specific to design patterns in socially interactive robots. Through the inspection of this set of features, designers from other areas of design can identify holes in the design space that have been addressed in other areas of design.

We noted some interesting trends in the design space from this coding process. For example, the heights of the robots in this database appears strongly bimodal, with one peak at robots near 25cm in height and the other peak at robots 150cm in height. The most common color of robot in out database was overwhelmingly white with 101 robots, followed by blue with 23 robots, and black with 13 robots. By manually describing embodiments in terms of design choices encompassed in their physical appearance, designers can evaluate how the combination of several design choices may affect different expectations. This coding process allows designers to develop new ways of describing embodiments as design patterns change over time as societal tastes change. 

% Some examples of general properties of the robots include height (M=88cm, SD=57cm), weight (M=38kg, SD=46kg), most prominent color (top three: 101 white, 23 blue, 13 black; expressed continuously as RGB), mechanically actuated face (present in 26 robots, absent in 141 robots), and number of legs (range=0-4 legs, median=0 legs).

\subsection{Visualizing the Design Space}
\begin{figure*}
     \centering
     \hspace*{\fill}%
     \begin{subfigure}[b]{0.45\textwidth}
         \centering
         \includegraphics[width=\textwidth]{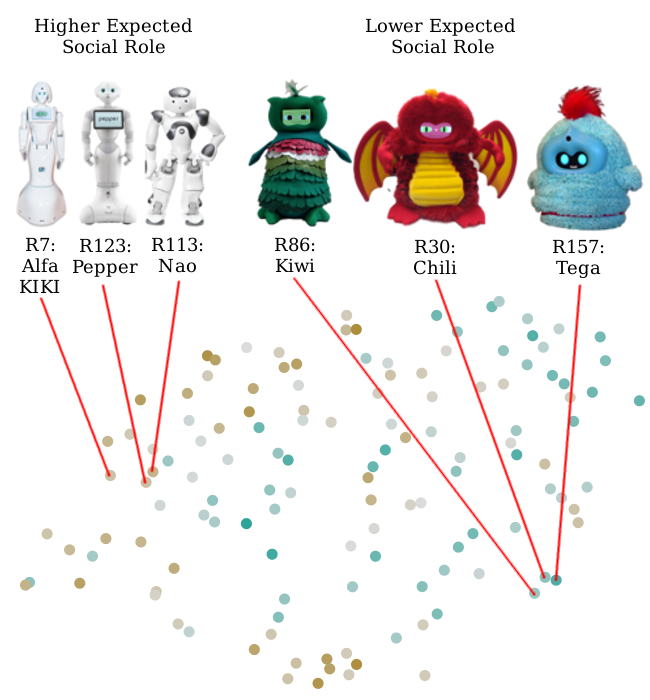}
         \caption{t-SNE plot of expected social role.}
         \label{fig:socialRole}
     \end{subfigure}
     \hfill
     \begin{subfigure}[b]{0.45\textwidth}
         \centering
         \includegraphics[width=\textwidth]{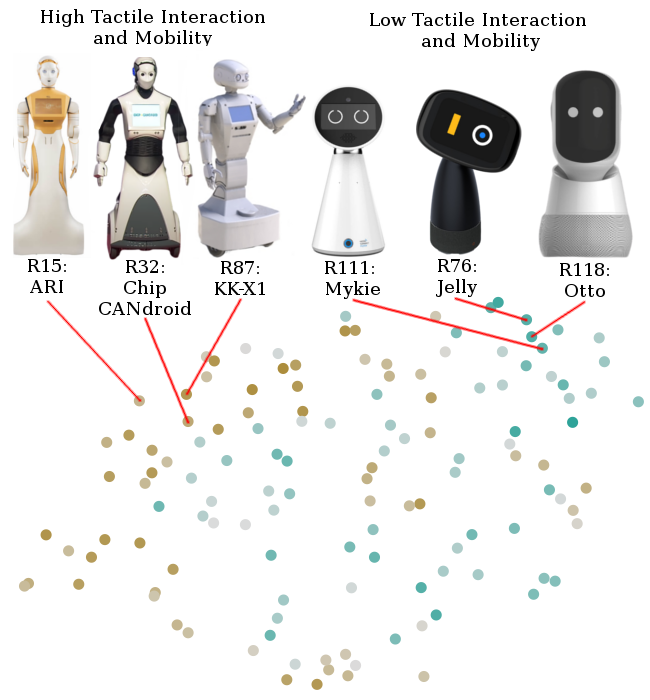}
         \caption{t-SNE plot of tactile interaction and mobility.}
         \label{fig:TI}
     \end{subfigure}
     \hspace*{\fill}%
     \caption{A t-SNE visualization of the design space of robot embodiments. Each point represents one robot in the dataset. Brown represents high values and teal represents lower values of the measured ratings.}
     \Description{Two scatterplots of points are shown on a color-blind safe and perceptually-linear diverging scale from teal to brown where teal represents low values and brown represents high values for a given construct. For each scatter plot, three robots are selected from nearby points with high or low values on the scatterplot. The three robots shown for higher expected social role are called Alfa KIKI, Pepper, and Nao. They have similar abstractly human looking embodiments as well as minimal faces and  white exteriors. The robots with lower expected social roles are called Kiwi, Chili, and Tega. All three are colorful robots that have fuzzy exteriors. The high tactile interaction and mobility robots are called ARI, Chip CANdroid, and KK-X1. All three are mounted on top of wheeled platforms, and have two arms and a screen chest. The low tactile mobility robots are Mykie, Jelly,and Otto. All three look like screens on a stand with minimalist rendered faces.}
     \label{fig:tsne}
\end{figure*}

To visualize the design space, we used the hand-crafted features developed in Section~\ref{datasetOverview} as descriptions of the physical attributes of the robots in our dataset. To learn a mapping without supervision from that high-dimensional feature space to 2D, we used t-Stochastic Neighbors Embedding \cite{maaten2008visualizing} that preserves distances between points from high-D to 2D space. Figure \ref{fig:tsne} demonstrates that robots mapped near each other share similar characteristics. We show evaluations of different robots with different color values in 2D space. Higher values are concentrated in different parts of the space, indicating differences in social and functional expectations of the robot embodiments. This visualization technique can be used as a design tool to rapidly explore different robot embodiments for a desired set of expectations related to specific tasks.

\begin{figure}[ht]
  \centering
  \includegraphics[width=.95\linewidth]{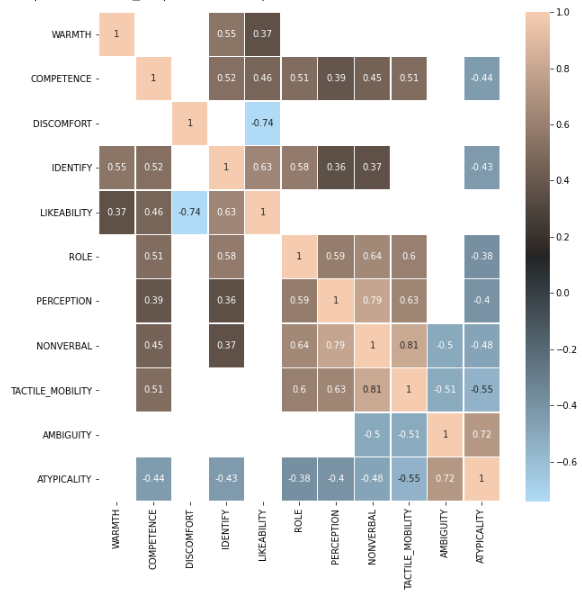}
  \caption{Correlations between attributes collected through surveys. All shown correlations are significant with $p < 0.001$. Pearson's r value is shown in each square. The first six items are from the survey of social perceptions, and the next five are from the survey of functional perceptions.}
  \Description{A heat map shows the values for all pairwise correlations of constructs that were measured in the surveys.}
   \label{fig:corr}
\end{figure}

\subsection{Correlations of Measures}
We show the correlations between all measures in Figure \ref{fig:corr}. While many correlations in the large collected dataset are significant, the coefficients of correlation are relatively small. We consider values of Pearson's $r>0.5$ to be of importance for reporting and discussion, representing a moderate correlation in similar contexts \cite{akoglu2018user}. The key findings are described in detail, with all reported correlations being significant with $p<.001$ after correcting for 110 pairwise comparisons of the 11 measures. All statistical analyses were performed with the Pinguoin library for Python \cite{vallat2018pingouin}. The following are the main trends in the results.

\subsubsection{Identity closeness is correlated with positive social perceptions.}
Identity closeness had a moderately strong correlation with warmth (r(163)=.55) and competence (r(163)=.52). Interestingly, the correlation with the discomfort construct is small. This is possibly related to the idea that discomfort is a construct that uniquely applies to robots, rather than warmth and competence, which apply to people as well as robots \cite{carpinella2017robotic}. Furthermore, the identity closeness of the user with the robot is strongly correlated with the likeability of the robot (r(163)=.63) and the expected social role of the robot (r(163)=.53). For the socially interactive robots we tested, we found that the closer the raters identified with the robot, the more positively they viewed the robot and the more likely the were to view the robot as a peer or superior.

\subsubsection{Likeability is not significantly correlated with perceived functionality.}
Likeability responses are moderately correlated with warmth (r(163)=.37) and competence (r(163)=.46), strongly anticorrelated with discomfort (r(163)=-.74), and strongly correlated with identity closeness (r(163)=.63). However, the reported likeability is not strongly correlated with any measures of perceived functional ability. In general, as raters felt more socially close to the robot and the design of the robot was less discomforting, users reported liking the robot more. Changes in perceived physical capabilities of the robot, however, did not correspond with a discernible change in how much users liked the robot.

\subsubsection{A robot's role is correlated with its functionality.}
The expected social role of the robot is strongly correlated with perception and interpretation (r(163)=.59), non-verbal communication (r(163)=.64), and tactile interaction and mobility (r(163)=.60). The social role is also correlated with the competence (r(163)=.51) and social identity (r(163)=.63). As functional abilities increased for robots in our dataset, raters were more likely to view them as peers or superiors. 

\subsubsection{The perceived functionalities of robots are entangled.}
For all pairwise comparisons of the functional constructs from the modified EmCorp-Scale, we observed correlation values larger than r(163)=.63. Thus, increases of one construct from this scale were associated with increases in the other two constructs of the scale for the set of socially interactive robots we tested. This implies that robots that appear more capable of moving through space \textit{additionally} elicit higher expectations of perceptual and interpretive abilities, as well as higher expectations of non-verbal expressivity than robots that do not appear as capable of moving through space. These three functional constructs can be interpreted together as a generic measure of the robot's holistic capability to interact with other agents and the world.  

\subsection{Predictive Features for Each Construct}
To relate the physical aspects of a robot's design to its expected social and functional affordances, we performed feature selection on the manually coded features for each of the measured quantitative perceptual constructs.  We used the Boruta algorithm \cite{kursa2010feature} because it aims to find \textit{all-relevant} features (i.e., all features that carry information on the modeled construct) as opposed to \textit{minimal-optimal} features (i.e., the minimum set of features that maximize predictive accuracy for some specific model).  The Boruta algorithm selects important attributes and is stable and unbiased when feature importance is measured with random forests of unbiased weak classifiers \cite{kursa2010feature}.  

We performed feature selection by creating "shadow features" of the true features by randomly permuting the true values. Both the true and shadow features are used to predict the value of a construct. If a true feature is given importance that is higher than its shadow feature, it is considered useful in classification. This process was run 500 times for statistical validity. Due to the exploratory nature of this work, we selected features that were more relevant than their shadow features with a probability of $0.5$. By selecting features that were relevant to specific constructs, we presented possible directions for investigating the relationship between robots' embodiments and their perceived expectations. Table \ref{table:selectedconstructs} shows the selected features as they relate to the measured constructs.

\begin{table}
\begin{tabular}{ |p{3cm}||p{9cm}|  }
\hline
 \textbf{Construct} &  \textbf{Relevant Selected Features (and Relationship)}\\
 \hline
 Warmth & \textit{Mouth?} (+) \\ \hline
 Competence&   \textit{Height} (+), \textit{Humanoid Embodiment?} (+)  \\ \hline
 Discomfort & \textit{Height} (+), \textit{Year} (-), \textit{Mechanical Face?} (+), \textit{Industry?} (-)\\ \hline
 Femininity &  \textit{Height} (-), \textit{Weight} (-), \textit{Most Prominent Color = Beige} (+), \\
 ~ & \textit{Blush?} (+), \textit{High Waist-Hip Ratio?} (+), \textit{Curved Embodiment?} (+) \\ \hline
 Identity Closeness & \textit{Height} (+), \textit{Humanoid Embodiment?} (+) \\ \hline
 Likeability &  \textit{Height} (-), \textit{Industry?} (+)\\ \hline
 Masculinity & \textit{ Height} (+), \textit{Weight} (+), \textit{Year} (-), \textit{Curved Embodiment} (-), \\
 ~ & \textit{Jointed Limbs?} (+)\\ \hline
 Social Role &  \textit{Height} (+), \textit{Year} (-), \textit{Humanoid Embodiment?} (+), \\
 ~ & \textit{Number of Arms} (+)\\ 
\hhline{|=||=|}
 Perception and  & \textit{Height} (+), \textit{Humanoid Embodiment?} (+), \\
 Interpretation & \textit{Dominant Classification}=Anthropomorphic (+)\\  \hline
 Tactile Interaction  &  \textit{Height} (+), \textit{Mobile?} (+), \textit{Number of Wheels} (+), \\
 and Mobility & \textit{Number of Arms} (+), \textit{Jointed Limbs?} (+)\\  \hline
 Nonverbal  &  \textit{Height} (+), \textit{Year} (-), \textit{Humanoid Embodiment?} (+), \\
  Communication & \textit{Number of Wheels} (-), \textit{Number of Legs} (+), \textit{Number of Arms} (+), \textit{Dominant Classification = Anthropomorphic} (+), \\
  ~ & \textit{Jointed Limbs?} (+)\\  \hline
 Design Ambiguity &  \textit{Height} (-), \textit{Weight} (-), \textit{Number of Legs} (-), \\
 ~ & \textit{Dominant Classification = Anthropomorphic} (-)\\  \hline
 Design Atypicality & \textit{ Weight} (-), \textit{Number of Legs} (-), \\
 ~ & \textit{Dominant Classification = Anthropomorphic} (-)\\
 \hline
\end{tabular}
 \caption{The important features as selected by the Boruta algorithm from our manually coded feature set that corresponded to the constructs measured in the survey.}
 \label{table:selectedconstructs}
\end{table}

These selected features can be separated into two categories: unobservable and observed. From the unobservable features we can analyze trends in the design space that are reflected in the overall design of robot embodiments as opposed to specific aspects of embodiments. The main unobservable features of importance were: \textit{Year}, the year of release of the embodiment and \textit{Industry?} whether or not the robot was at one point commercially available. The year descriptor allowed us to capture the non-stationary nature of design practices over time. Most notably, newer robots in our database were in general less discomforting, less stereotypically masculine, had a lower expected social role, and had fewer non-verbal communicative capabilities. The commercially available attribute is related to the effect that larger teams of designers have on the development of robotic systems. Robots that were commercially available were more likely to be more likeable and less discomforting. This suggests that robots used in settings that require the robot to be a comforting partner in interaction may benefit from using robots that are commercially available rather than robots developed for research purposes.

Of the observed features of importance, height was the most frequently selected feature across all of the measured constructs. Height has previously been related to increased expected social role \cite{rae2013influence} in controlled settings, height as an important characteristic in this in-the-wild setting suggests that this finding generalizes beyond the laboratory setting. However, height's relationship with other constructs has not been studied closely, opening up interesting research questions for future work.

Another trend we noted is the importance of anthropomorphism in functional constructs. In general, robots that are seen as human-like are expected to have higher degrees of functional capability, as well as to take on more superior social roles. This elevated expectation, however, requires that the robot actually meet these expectations. Thus, when using anthropomorphic embodiments, care should be taken to ensure that the robots operate to their expectation.

Our results replicated findings that related body shape to the expression of femininity in robots. Previous work has similarly linked the relationship between robots' waist-to-hip ratio to their perceived gender expression \cite{bernotat2017shape, trovato2018she}. Similarly, \citet{kalegina2018characterizing} found a relationship between perceived gender and the presence of blush. These features, however, are not important in predicting the masculinity ratings of the robots. This suggests that the axes of femininity and masculinity in robots are not diametrically opposed.

\section{Utility of Metaphors}
\subsection{Metaphor Summary}
\begin{figure*}[t]
  \centering
  \includegraphics[width=\linewidth]{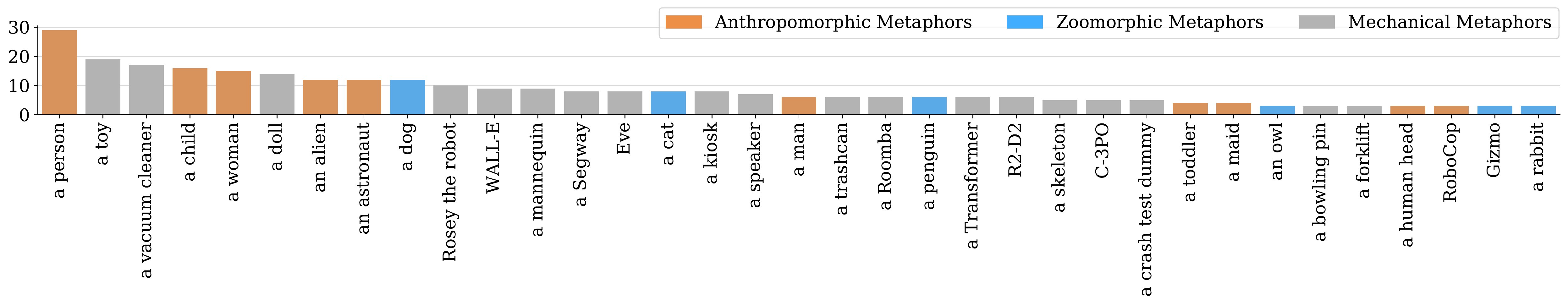}
  \includegraphics[width=\linewidth]{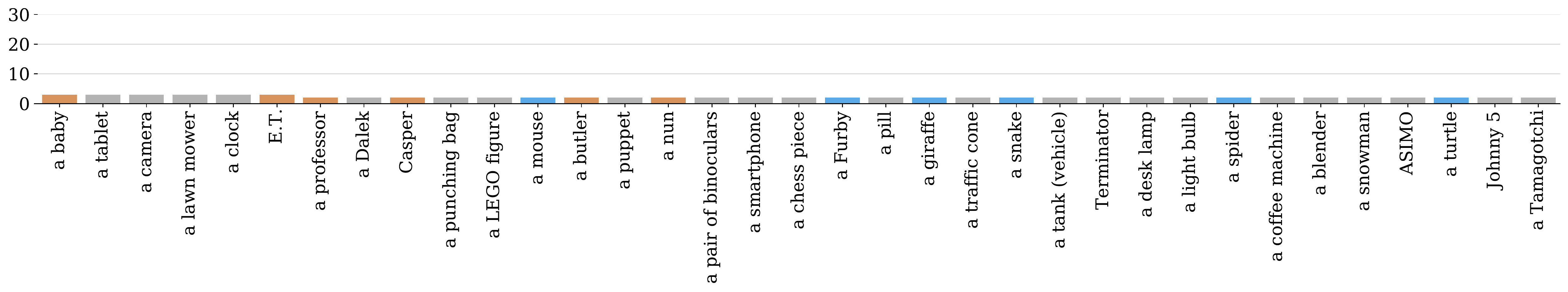}
  \caption{Histogram of metaphor counts for metaphors that were associated with more than one robot in our dataset. An additional 129 singleton design metaphors are not shown. The specific sources of the referential metaphors (e.g., movies, shows) are removed for visualization, but were provided to participants during data collection.}
  \Description{A histogram of design metaphors, approximating an exponential distribution. The metaphors in decreasing order are "a person", "a toy","a vacuum cleaner", "a child", "a woman", etc.}
   \label{fig:metaphorDist}
\end{figure*}

The participants' metaphors were unconstrained in the data collection, and the frequency of metaphors approximated an exponential distribution, as shown in Figure \ref{fig:metaphorDist}, with several metaphors appearing repeatedly throughout the dataset.

In total, 199 metaphors were used by the participants to describe all the robots in the database. We classified these metaphors as: anthropomorphic, zoomorphic, or mechanical. Metaphors were sorted by their literal interpretations; all nonliving metaphors were considered mechanical, living metaphors that represent animals were considered zoomorphic, and living metaphors representing humanoids were considered anthropomorphic. Of the 199 metaphors, 46 were classified as anthropomorphic, 46 were zoomorphic, and 107 were mechanical. We additionally observed that 38 of the metaphors were references to robots from popular media, such as Disney's WALL-E and The Jetsons' Rosey the Robot.

\subsection{Robot Metaphor Category Ascription and Manipulation Verification}
To quantitatively evaluate the differences in robot perception between metaphors, we assigned each design metaphor to one of the following groups: anthropomorphic, zoomorphic, or mechanical, as in previous studies \cite{kalegina2018characterizing, li2010cross}. Embodiments were assigned to the three categories based on the majority of the assignments of the top three metaphors. For robots with one of each kind of metaphor, we chose the metaphor with the highest number of responses. Based on these criteria, the database consistent of 46 anthropomorphic, 28 zoomorphic, and 91 mechanical robots.

To verify that these groups are meaningful, we performed a manipulation check with the open-source ABOT database \cite{phillips2018human}, a collection of similar robots that are rated on their human-likeness. We selected the robots that occurred in both databases and compared their human-likeness to verify that the assignment of those categories are meaningful. The intersection of the two databases contained 44 mechanical, 36 anthropomorphic, and 6 zoomorphic robots. Using a Welch's ANOVA test for unequal group sizes, we found that robots with both mechanical metaphors ($M_{human-likeness} = 23.58$) and zoomorphic metaphors ($M_{human-likeness} = 30.61$) were significantly less human-like than robots with anthropomorphic metaphors ($M_{human-likeness} = 47.87$), with Welch's F(2,14.60) = 17.92, $p<.001$, $\eta^2 =.33$. This affirms that our assignment of robots to metaphor types is not arbitrary.

\subsection{Differences Between Metaphor Categories}\label{metaphorCategories}
\begin{figure*}[t]
  \centering
  \includegraphics[width=\linewidth]{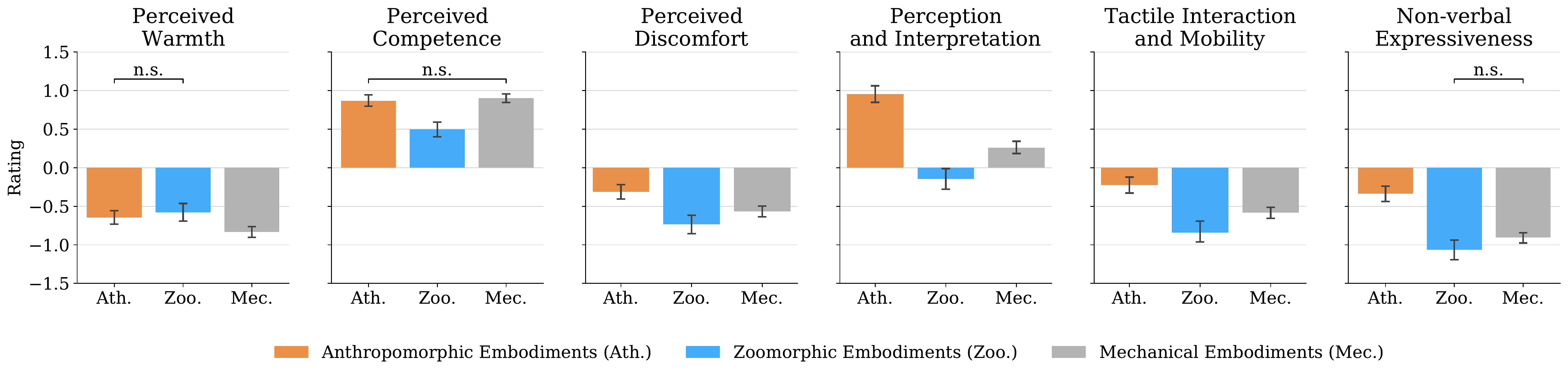}
  \caption{Differences in means for anthropomorphic, zoomorphic, and mechanical embodiments for social and functional constructs of embodiment. All differences are significant with $p < 0.001$ unless marked otherwise. Error bars represent 95\% CI of means.}
  \Description{The six graphs show the means of the different forms of embodiment (anthropomorphic, zoomorphic, and mechanical) for the six main constructs that were measured (Perceived Warmth, Perceived Competence, Perceived Discomfort, Perception and Interpretation, Tactile Interaction, and Mobility).}
   \label{fig:betweenmetaphors}
\end{figure*}

 We found differences in mean construct ratings for each classification of metaphor via a Welch's ANOVA tests, due to the unequal group sizes and heteroscedasticity of variances between groups. For post hoc analysis, we used the Games-Howell post hoc test to test pairwise comparisons between groups of unequal size and different variances, with results shown in Figure \ref{fig:betweenmetaphors}.

\subsubsection{Metaphor Type and Social Expectation} For the socially interactive robots we tested, the different categorizations of robot types had significant effects on participants' social expectations. For warmth, the main effect of group type was significant, Welch’s F(2,  1382.94) = 9.28, $p < .001$, $\eta_p^2=.005$. Post hoc analysis revealed that the mean warmth of anthropomorphic embodiments (M=-0.58) was significantly higher than the mean warmth of mechanical embodiments (M=-.83), $p=.001$, $\eta^2=.007$, and the mean warmth of zoomorphic embodiments (M=-.65) was significantly higher than that of mechanical embodiments, $p=.003$, $\eta^2=.004$.

The main effect of group type for competence was also significant with Welch’s F(2, 1353.95) = 24.09, $p < .001$, $\eta_p^2=.016$. The difference between perceived competence in anthropomorphic embodiments (M=.86) was significantly higher than the mean competence of zoomorphic embodiments (M=.50), $p=.001$, $\eta^2=.022$,  and the  mean competence of mechanical embodiments (M=.90) was significantly higher than the mean competence of zoomorphic embodiments, $p=.001$, $\eta^2=.026$.

Significant differences in discomfort were also observed across robot types, with Welch’s F(2, 1399.77) = 16.13, $p < .001$, $\eta_p^2=.010$. Zoomorphic embodiments (M=-.73) were rated significantly lower in discomfort than mechanical embodiments (M=-.57), $p=.04$, $\eta^2=.003$, followed by anthropomorphic embodiments (M=-.31), $p=.001$, $\eta^2=.007$, and zoomorphic embodiments were rated significantly lower in discomfort than anthropomorphic embodiments, $p=.001$, $\eta^2=.042$.

\subsubsection{Metaphor Type and Functional Expectation} The different categorizations of metaphor type additionally had effects on how participants expected a robot to perceive and interpret the world, with Welch’s F(2,  1297.52) = 94.36, $p < .001$, $\eta_p^2=.053$. Post hoc analysis revealed significance between all pairwise comparisons with $p<.001$. Zoomorphic embodiments were perceived as having the lowest perceptual capabilities (M=-.14), followed by mechanical embodiments (M=.26), $\eta^2=.018$, and then followed by anthropomorphic embodiments (M=.95), $\eta^2=.046$. Therefore, anthropomorphic embodiments had a much larger difference in perceived perceptual abilities than zoomorphic embodiments, $\eta^2=.118$.

Tactile interaction and mobility also showed differences in metaphor types with Welch’s F(2,  1209.25) = 27.81, $p < .001$, $\eta_p^2=.019$. All pairwise comparisons were significant in the post hoc analysis with $p<.001$. Zoomorphic embodiments were perceived as having the lowest ability to manipulate objects in the world (M=.84), followed by mechanical embodiments (M=-.58), $\eta^2=.007$, and then followed by anthropomorphic embodiments (M=-.22), $\eta^2=.013$. Zoomorphic embodiments therefore had much lower perceived tactile abilities than anthropomorphic embodiments, $\eta^2=.037$.

The different forms of embodiment showed different expectations to communicate non-verbally with Welch’s F(2, 1239.08) = 49.65, $p < .001$, $\eta_p^2=.032$. Zoomorphic embodiments (M=-1.06) were perceived as less capable of communicating non-verbally than anthropomorphic embodiments (M=-.33), $p=.001$, $\eta^2=.055$. Mechanical embodiments (M=-.91) were also viewed as having lower non-verbal communicative abilities than zoomorphic embodiments, $p=.001$, $\eta^2=.033$.

\begin{figure}[t]
  \centering
  \includegraphics[width=.7\linewidth]{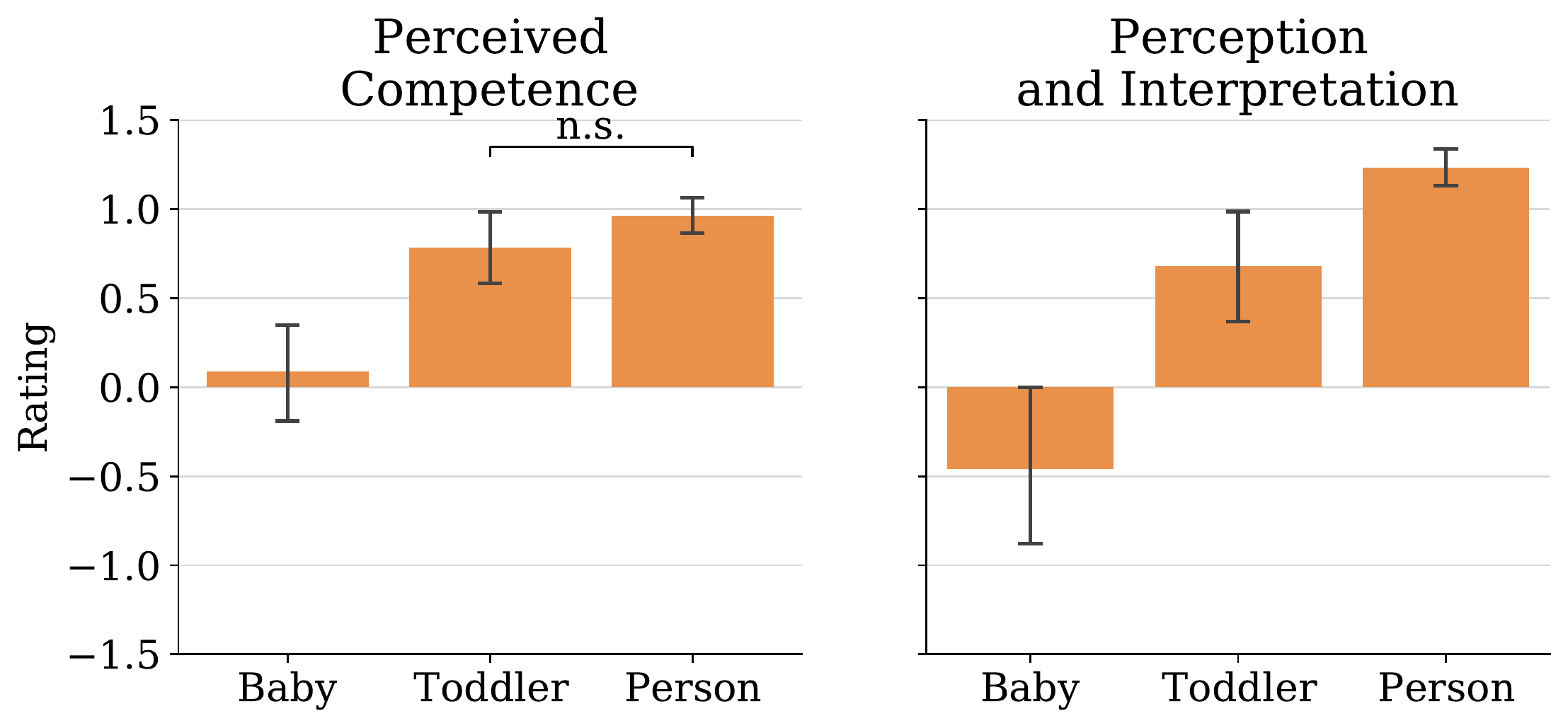}
  \caption{Perceived competence and perceived perceptual ability by metaphors for different maturity levels.}
  \Description{Bar graphs showing the mean of rating for perceived competence and perception and interpretation as a metaphor progresses by maturity. The progression of metaphors is "baby", "toddler" and "person". Perceived competence increases along this progression, as does Perception and Interpretation Abilities.}
   \label{fig:competenceByMaturity}
\end{figure}

\subsection{Design Metaphor Semantics}
We were also interested in exploring whether design metaphors are semantically meaningful in terms of user perceptions. While the semantic space of metaphors is difficult to describe, there are some locally ordered areas. To examine the effects of social and functional perceptions, we selected three metaphors: "a baby", "a toddler", and "a person". Because age is associated with competence \cite{khadpe2020conceptual} and interpretation of the world, we expected that robots described with more mature metaphors would have higher competence and perceptual capabilities.

As expected, we found that a main effect was present on metaphor name and competence with Welch’s F(2,  127.81) = 16.55, $p < .001$, $\eta_p^2=.057$. The perceived competence is lower for robots labeled with the metaphor "a baby" (M=.08), followed by robots described with the metaphor "a toddler" (M=.78), $p = .001$, $\eta^2=.09$, and then followed by robots described with the metaphor "a person" (M=.96), $p= .001$, $\eta_p^2=.110$.

There was an additional effect on the perceived perceptual abilities of robots with Welch’s F(2,  96.60) = 30.81, $p = .001$, $\eta_p^2=.120$. Robots described as babies were assumed to have lower expected perceptual capabilities (M=-.45) than robots described as toddlers (M=.67), $p = .001$, $\eta_p^2=.118$. Robots associated with the toddler metaphor were, in turn, perceived as having lower perceptual abilities than robots described as persons (M=1.23), $p = .001$, $\eta_p^2=.110$. Additionally, robots associated with the baby metaphor had significantly lower perceived perceptual abilities than robots associated with a person metaphor, $p = .001$, $\eta_p^2=.212$.

\section{Robots as Social Actors} \label{socialExpectations}
In addition to the overall trends, we aimed to investigate two other major areas of social expectations that are influenced by robot embodiment: (1) the expression of robot gender \cite{eyssel2012s,trovato2018she,winkle2021boosting} and (2) the formation of social group memberships with robots \cite{eyssel2012social,fraune2017teammates, fraune2020some}. We sought to understand how metaphor attribution may affect these two social constructs.

\subsection{The Space of Robot Gender Expression}

\begin{figure*}[t]
     \centering
     \hspace*{\fill}%
     \begin{subfigure}[b]{0.32\textwidth}
         \centering
         \includegraphics[width=\textwidth]{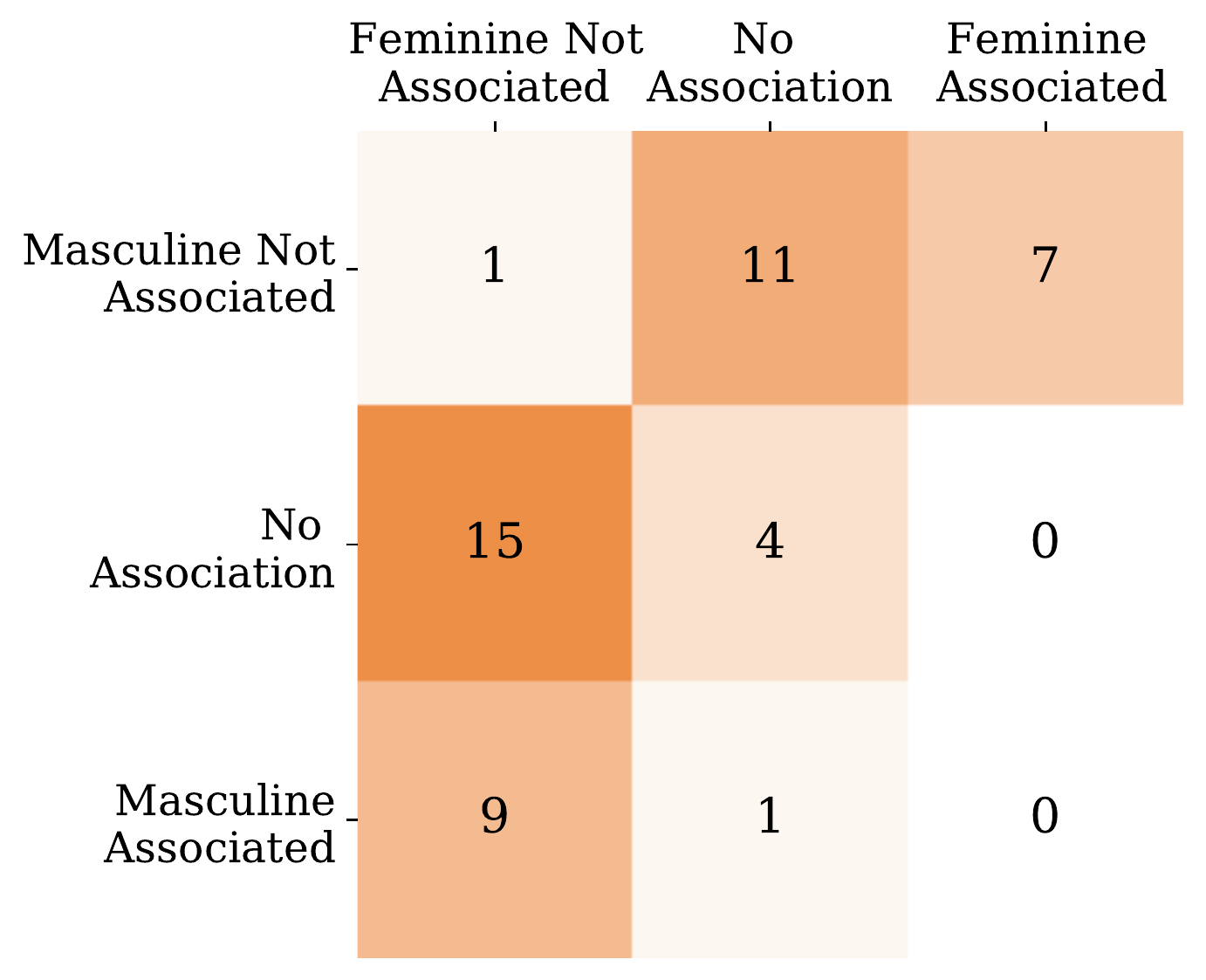}
         \caption{Anthropomorphic embodiments.}
         \label{fig:anthropomorphicGender}
     \end{subfigure}
     \hfill
     \begin{subfigure}[b]{0.32\textwidth}
         \centering
         \includegraphics[width=\textwidth]{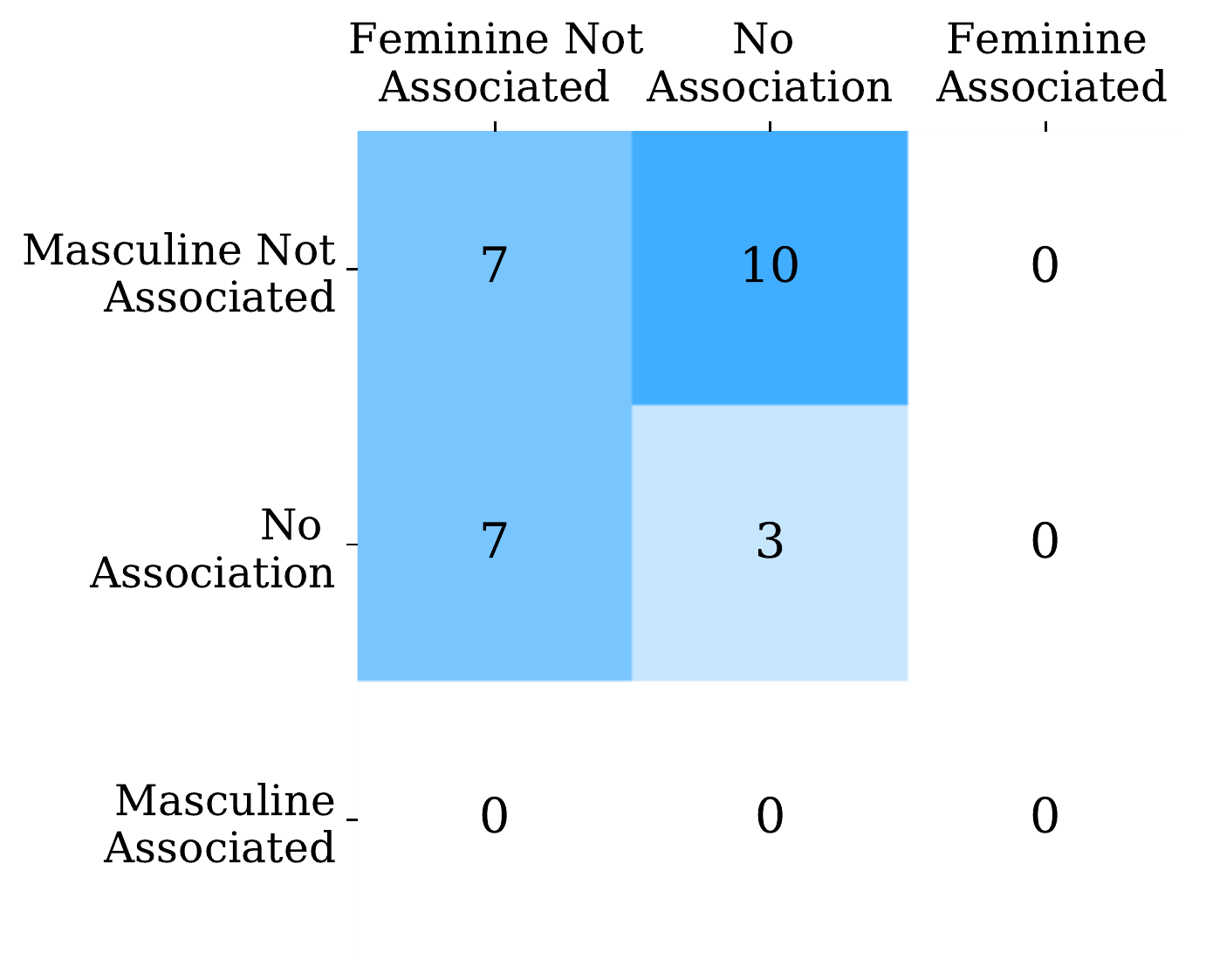}
         \caption{Zoomorphic embodiments.}
         \label{fig:zoomorphicGende}
     \end{subfigure}
     \hfill
     \begin{subfigure}[b]{0.32\textwidth}
         \centering
         \includegraphics[width=\textwidth]{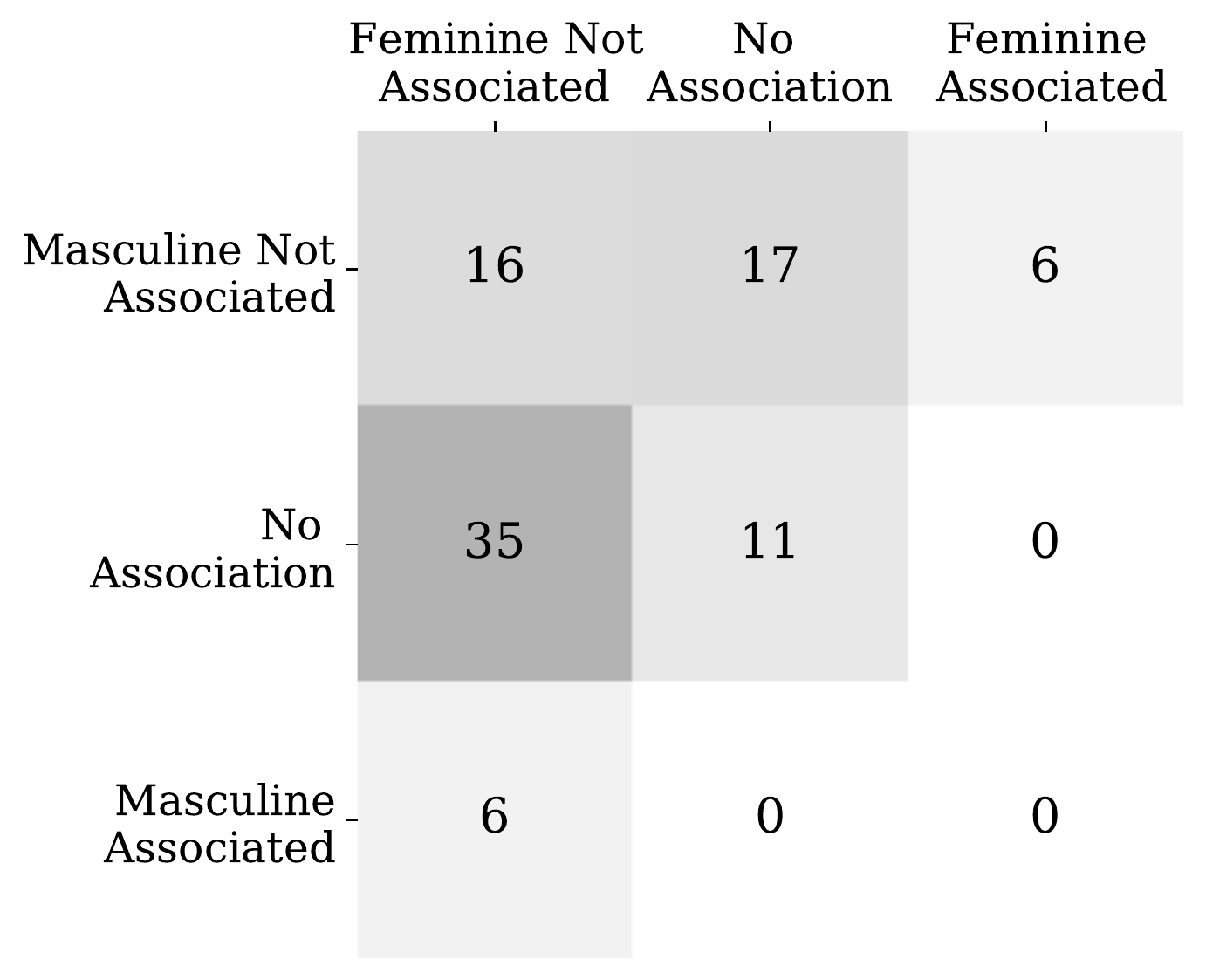}
         \caption{Mechanical embodiments.}
         \label{fig:mechnicalGender}
     \end{subfigure}
     \hspace*{\fill}%
     \caption{A visualization of the space of gender expression by robot metaphor type.}
     \Description{}
     \label{fig:genderSpace}
\end{figure*}

To examine the space of gender expression (i.e., how masculinity and femininity are embodied \cite{anderson2020gender}; see Section~\ref{study2measures} regarding the selection of these axes), we constructed the space according to the results of a two-tailed Wilcoxon signed-rank test. For each robot, we independently determined if the robot's average ratings for femininity and masculinity were significantly above zero, significantly below zero, or the null hypothesis that the value is zero could not be rejected. A value of zero corresponded to masculinity or femininity being neither associated with the robot nor not associated with the robot. Approximately, the cutoffs for the robots were around average values of $\pm 1$, corresponding to "slightly agree" and "slightly disagree". The results of this investigation are shown in Figure \ref{fig:genderSpace}.

By separating across design metaphor classification, we observed patterns in the way that the gender expressions were perceived. Anthropomorphic embodiments were more likely to be perceived as having a significant association with either femininity or masculinity. Zoomorphic robots were unlikely to be associated with a particular axis of gender expression. Mechanical embodiments were more likely to have no gender expression association, but in some cases they were associated with either masculinity or femininity.

These findings highlight areas of the design space that merit further exploration. There were relatively few robots in the database that portrayed both masculine and feminine characteristics of gender expression. One anthropomorphic robot did exhibit masculine characteristics and some degree of feminine characteristics. Interestingly, we also observed that anthropomorphic embodiments and mechanical embodiments skewed toward not being associated with femininity, while zoomorphic embodiments skewed slightly toward not being associated with masculinity. This suggests future directions of design research that could explore how to balance these trends in the current design space.

\subsection{Formation of Ingroups and Outgroups}

To evaluate the formation of social group membership, we used the measure of identity closeness. We observed that many of the robots in the database experienced bimodal distributions, indicating that the formation of ingroups and outgroups may occur based on the robots' embodiments. Responses to Likert scale questions have been shown to follow binomial distributions in past work \cite{allik2014mixed}. To evaluate this possibility in our data, we modeled the responses as coming from two possible models: a unimodal binomial model and a bimodal binomial model, with priors as described below. All models were developed and fit to the observed data using the pymc3 framework \cite{salvatier2016}.  The following describes the unimodal model:

\begin{gather*}
    p \sim Beta(1,1)\\
    Y \sim Binomial(p,N)
\end{gather*}

where $p$ represents the probability of success of a binomial trial with N repetitions. We used a 9-point Likert scale, thus N was set to 8. The prior for $p$ was characterized as an uninformative Beta distribution. Y represented the values that users responded with.  The following describes the bimodal model:

\begin{gather*}
    p_{ingroup} \sim Beta(2,1)\\
    p_{outgroup} \sim Beta(1,2)\\
    w \sim Dirichlet(1,1)\\
    Y \sim \sum_{i\in \{ingroup, outgroup\}} w_i \cdot Binomial(p_i,N)
\end{gather*}

The bimodal model is a weighted sum of two binomial distributions characterized with two different probabilities of binomial trial success, $p_{ingroup}$ and $p_{outgroup}$, which have weights corresponding to $w$. The ingroup distribution and outgroup distribution had equal but opposite uninformative priors to ensure stability across different threads of MCMC sampling, since prior research has shown that ingroup membership is related to being closer than outgroup membership \cite{tajfel1974social,reysen2013further}. All models were evaluated with two independent sampling chains with 20,000 iterations to guarantee convergence to the observed posterior distribution.

The binomial model was selected over the unimodal model if it was more than 10 times more likely based on the Watanabe Aikiake Information Criterion (WAIC) for the observed data. We found that of the 166 total robots, 60 were best described with the unimodel model of group membership, and 106 were best described with the bimodal model of group membership. A Chi Square test revealed that the distributions of metaphor types within these groups were significantly different from each other, $\chi^2(2, N=166) = 59.33, p<.001$, with anthropomorphic and mechanical metaphors being more represented by unimodal model and zoomorphic robots being more often represented by the bimodal model. This suggests that robots were likely to form ingroups and outgroups based on their design. 

Our results extend work that has identified the effect of group membership by developing robot identity through personality \cite{eyssel2012social} to include robot identity defined through its embodiment. They also extend work that focused on group formation based on robot color \cite{kuchenbrandt2013robot} to encompass holistic morphologies.

\section{Robots as Functional Agents} \label{functionalExpectations}
In addition to social expectations discussed so far, functional expectations are also set by the robot's embodiment. To understand how, we investigated the tasks that study participants assigned to each robot using a grounded theory approach. We additionally examined specific metaphors in detail to understand the effects of functional expectations based on how strongly the metaphor is evoked in the robot.

\subsection{Embodiment and Task}

\begin{table}[t]
\begin{tabular}{ |c c|c| }
\hline
\multicolumn{2}{|c}{\textbf{Assigned Task}} & \multicolumn{1}{|c|}{\textbf{Specific Population}} \\
 \hline
 Companion & Home Assistant & Children \\
 Customer Service & Informant & Elderly \\
 Educator &  Manufacturer & Persons with Disabilities\\
 Entertainer & Surveillant & ~\\
\hline
\end{tabular}
 \caption{A table of codes developed through qualitative analysis of the user-reported tasks for socially interactive robot embodiments.}
 \label{table:codes}
\end{table}

To evaluate the task expectations of the different robot embodiments in the database, we developed a coding scheme from the participants' free-response answers to the question regarding what task the robot appeared to be useful for. We observed both task-related and intended population remarks from the participants. Eight main task-related codes were developed and three specific population labels were identified as trends in the design space of socially interactive robots, as summarized in Table \ref{table:codes}. Interestingly, these codes have considerable alignment with the task categorization used by \citet{kalegina2018characterizing}, despite being collected in an open format. The key differences we found are that we did not observe high numbers of responses for performing research, nor for health-related tasks. In addition, we observed two additional categories: being used as a companion and being used to collect or provide information.

\begin{figure*}[th]
     \centering
    \includegraphics[width=\textwidth]{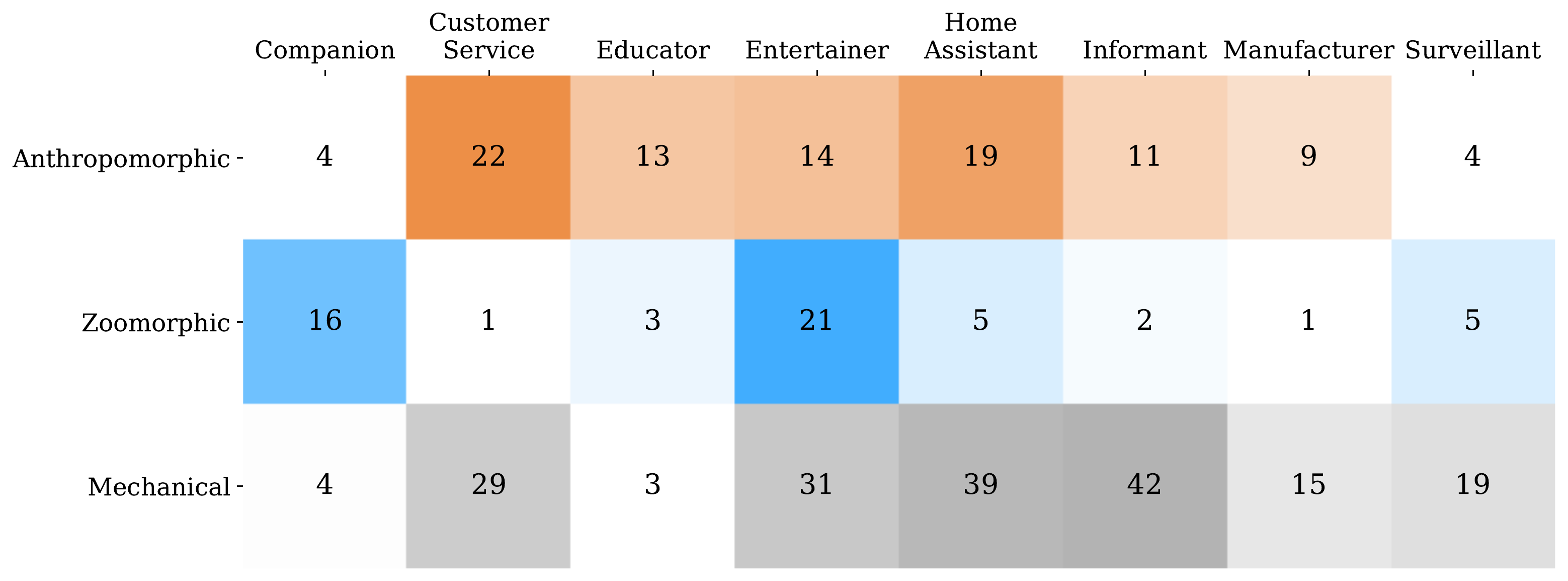}
    \caption{A map of the distribution of the top two tasks for robots in our dataset separated by their metaphor type.}
    \Description{}
    \label{fig:jobs}
\end{figure*}

The \textit{companion} context is characterized by tasks involving the robot acting socially to improve mood or mental health over long periods. Examples of common tasks for this context were robots that "provide warmth and comfort", "are an interactive friend for my child", and "being a conversation partner". Most commonly, zoomorphic robots were described as being appropriate for this task. This aligns with the zoomorphic robots' tendency to be perceived as comforting and warm, a key component of these tasks where functional expectation are not as important.

Robots ascribed for \textit{customer service} contexts were expected to directly interact with people in public places such as stores, restaurants, or hotels. Example tasks were robots that function as a "greeter or a receptionist", "a waiter" and "a museum guide". Both anthropomorphic and mechanical embodiments were described as being useful for customer service-type tasks. This aligns with the high expected functionalities of these embodiments to perform the services those tasks require.

\textit{Educator} tasks involved knowledge transfer from or through the robot to a person interacting with the robot. Tasks fitting this category involved robots that could be used "in language education", "to interact with students in class", and to provide "light educational lessons like spelling or math". Interestingly, the embodiment of the robot was often related to the topic that the robot was meant to teach. For example, the baby-like robot Babyloid was described as a "a training baby for expecting mothers", and the cat-like robot MarsCat could be used "to help educate about cats". Most commonly anthropomorphic embodiments were assigned to tasks relating to that topic. This is consistent with the high perceived competence and functionality of anthropomorphic embodiments.

For robots that played the role of \textit{entertainers}, expected tasks aligned with short-term entertainment purposes. For example, robots in this category were expected to "play music", "be used like a toy", and "tell jokes". This category was common across all types of metaphors, however each metaphor was described as entertaining in a specific way. Anthropomorphic metaphors were described as being used as "a game-playing partner", zoomorphic metaphors were most often seen as functioning like "a pet that doesn't require attention when not in use", and mechanical metaphors fulfilled roles that are common in other forms of technology such as "playing music".

\textit{Home assistant} robots were expected to work within the household, performing chores and other daily tasks, including "cleaning up after kids", "making coffee", and "carrying groceries". These tasks are similar to the customer service task, but are distinct in that they occur in the home and consist of repeated interaction with a few people. Similar to customer service tasks, both mechanical and anthropomorphic metaphors were well-suited for the home assistant task.

Robots that act as \textit{informants} were described with tasks that answer questions or otherwise provide information. Common tasks in this category were robots that "verbally answer questions", "tell time", or "report daily events like news or weather". Mechanical robots were most frequently described as being useful for these impersonal and intellectual tasks, consistent with their perceived high competence.

\textit{Manufacturer} robots were used in contexts where they build or move objects, typically without constant direct human interaction. These robots were expected to "carry heavy objects", "be a factory worker", and "pack in a warehouse". Mechanical embodiments and, to some extent, anthropomorphic embodiments, were selected for tasks like these, primarily for their functional capabilities, as these tasks were parceived to not require social interaction.

Robots that were perceived as \textit{surveillants} were those that monitor behavior, and were typically expected to provide security in some way. These robots were expected to be similar to "security alarms", "spy cameras", or "a sentry". Mechanical embodiments were most frequently attributed to this task. Similar to informants, these types of tasks are impersonal but require high levels of competence and perceptual capabilities, qualities attributed to mechanical embodiments.

\subsection{Abstraction and Functionality}
\begin{figure*}
     \centering
    \includegraphics[width=\textwidth]{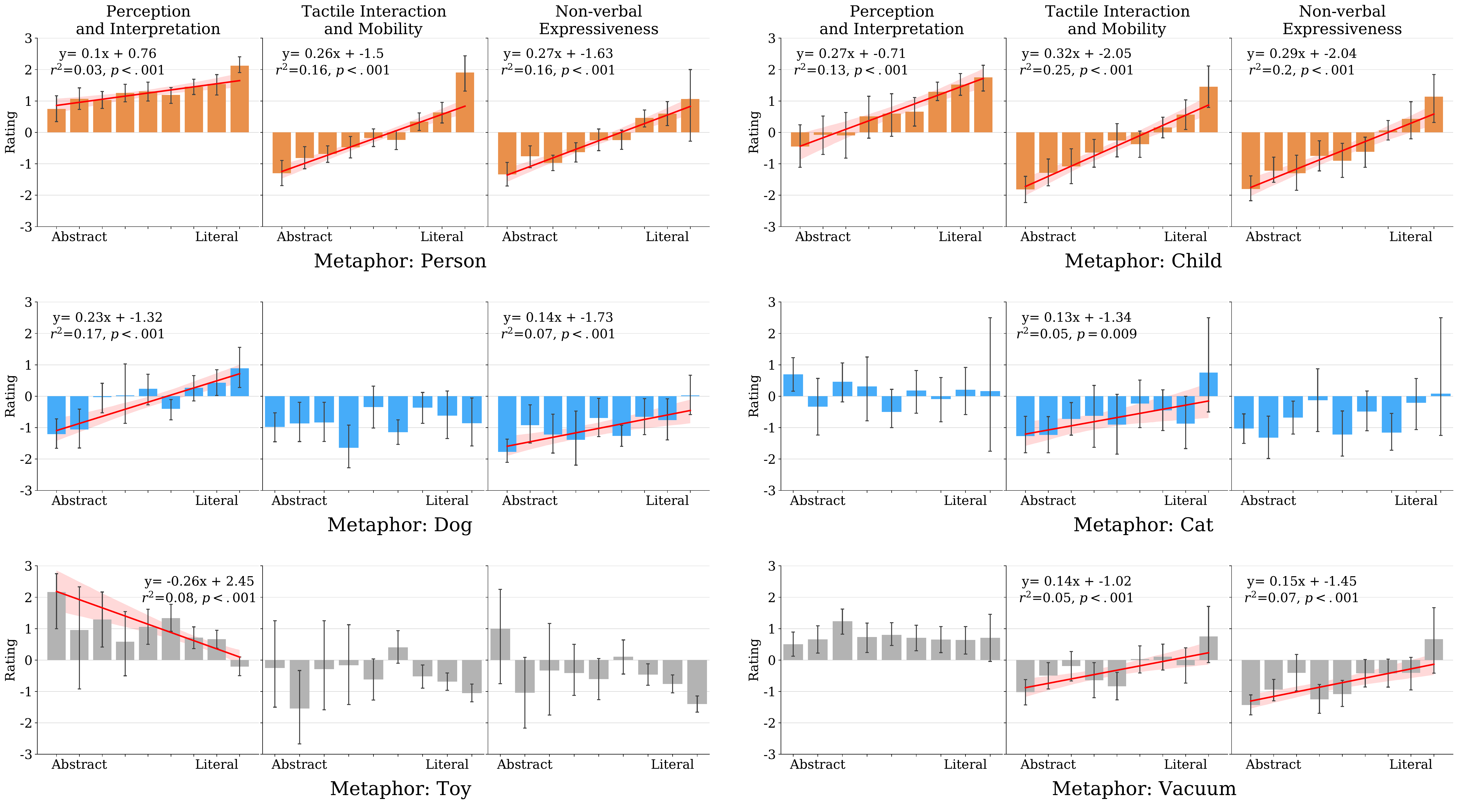}
    \caption{Plots showing the effect of metaphor abstraction on perceived functional characteristics of robots. Regression lines are shown for significant effects, after Bonferroni correction for 18 hypotheses.}
    \Description{}
    \label{fig:abstraction}
\end{figure*}

To investigate the level of abstraction of a metaphor, we selected the top two most frequent metaphors from each category. For anthropomorphic metaphors, the two were "a person" and "a child"; for zoomorphic metaphors they were "a dog" and "a cat"; and for mechanical metaphors, they were "a toy" and "a vacuum". For all metaphors, the rating of functional expectations were regressed onto the level of abstraction of the given metaphor. Significant regressions are shown in Figure~\ref{fig:abstraction}. The equation of the regression line is given along with the corresponding $r^2$ value.

For all anthropomorphic metaphors, we observed a significant increase in all perceived functional constructs as the robots were seen as more literal humans. This reflects similar findings from Section~\ref{metaphorCategories}, where anthropomorphic metaphors were consistently rated as having the highest functional expectation. This trend of increasing functional expectation as embodiments were perceived more literally across all constructs was shown in most of the anthropomorphic metaphors that we measured. This finding aligns with the idea of anthropomorphization of robots as assigning more human-like abilities to these embodiments, not only socially but functionally as well~\cite{duffy2003anthropomorphism}.

For zoomorphic metaphors, we observed different trends across metaphors. For dog-like robots, the perception and interpretation and the non-verbal expressiveness constructs significantly increased as the robots appeared more like real dogs. For cat-like robots, however, only tactile interaction and mobility construct increased with increasingly literally perceived implementations. This difference may be a result of commonly held views about these animals in the United States; dogs are typically seen as more attentive to their owners and are non-verbally expressive through bodily modalities such as tail-wagging. Cats, on the other hand, are seen as more passive in interaction but also as more dexterous.

Mechanical metaphors also exhibited different trends across metaphors in terms of their level of abstraction. Robots described as "a toy" were significantly \textit{less} perceptive and interpretive as they looked more like literal toys. A similar phenomenon was observed by \citet{hegel2008understanding}, where users reported a Lego robot as looking like a "toy" and perceived it as simply pushing buttons, whereas they described computers as performing calculations. Vacuum-like robots instead showed an increase in perceived tactile interaction and mobility as well as non-verbal communication capabilities as they appeared more like vacuums. This is likely because vacuums typically move around rooms as part of the cleaning process, and robot vacuums prevalent in popular culture (e.g., the iRobot Roomba) exhibit mechanistic non-verbal communicative abilities.

\section{Discussion} \label{discussion}
The analysis of the large dataset resulting from the three studies conducted in this work demonstrates many nuances of the design space of socially interactive robot embodiments. The results include multiple insights that can be used to inform design processes for various robotics contexts. Specifically, this work suggests how measuring social and functional attributes of embodiment via design metaphors can be used to estimate and evaluate designs of robot embodiments, and provides a methodology to situate novel embodiments in the space of extant embodiments. These tools inform the development of future socially interactive robots for effective human-robot interaction.

\subsection{Implications for Study Designers}
From a research perspective, design metaphors can be used to align the specific robot being used in an interaction with the affordances the interaction dictates. By using the database we developed, a set of candidate robots can be selected for a given interaction based on the similarity to the robot's expected task contexts. By considering several options, a practitioner can be informed about the embodiment choice for a specific context. Similarly, depending on the robots that a study designer has access to, minor modifications can be made to make the selected robot fit a given task.

In general, we find that the high-level metaphor groups are particularly important to consider when designing an interaction. Anthropomorphic robots are best suited for highly-functional tasks, however care must be taken to ensure that these robots perform to their expectation. Zoomorphic robots are generally perceived as the least functional, but are rated much higher in warmth and comfort. Thus, designs that are described like animals are most appropriate for tasks that focus on providing support rather than manipulating objects or moving through space. Mechanical designs are the least warm, despite high perceptual capabilities and competence. Mechanical metaphors may best be used in more impersonal contexts where the robot is viewed a tool to accomplish objective non-social tasks.

\subsection{Implications for Robot Designers}
For research in robot design, crowd-sourcing of design metaphors through the questionnaires used in this work provides a process to evaluate novel designs. This quantification of the design space presents the trends and voids in the design space. This understanding can help to direct the flow of design research toward new or under-explored areas by partitioning the space of socially interactive robots in appropriate ways: by metaphor, by gender expression, or by task expectation. By such partitioning of the design space, the contributions of specific designs can be explained in terms of their contributions to the space of socially interactive robot designs. By summarizing design findings in this form of intermediate-level design representation, specific design artifacts can be readily found and used to inform the design of socially interactive robots.

\subsection{Limitations and Future Work}
A key limitation of this work is the use of images to convey representations of inherently 3D real-world objects. Multiple views of the robots were shown in an attempt to mitigate this, but a 2D screen cannot fully reconstruct the impression of the 3D real-world embodiments. Additionally, social and physical contexts are not considered in this work. Contextual information can have great impact on how a user expects a robot to behave (e.g., \cite{banerjee2018effects, nigam2015social}).  Since this was a Mechanical Turk study, users did not interact with the robots. We cannot, therefore, infer how user expectation may be altered through real-world interactions and over time. These findings are best viewed as \textit{priors} on robot expectation before interaction occurs. 

We restricted our participants to the United States. Because many metaphors and perceptions may be culturally situated, it is not certain how these results may generalize to other cultures.  This issue could be addressed by following a similar design in future examinations of design metaphors in different cultural contexts.

This work utilizes the ontology of anthropomorphic, zoomorphic, and mechanical metaphors as a means of analysis. While that was a useful classification for our analysis, as the space of robot design expands and other metaphors are used in design processes, that ontology should be reconsidered to overcome its limitations (for example to capture robots that look like plants). Importantly, the utility of design metaphors as a tool for understanding user expectations does not directly depend on this ontology, allowing for the evolution of classification systems.

\section{Conclusion} \label{conclusion}
This work aims to provide a framework for understanding and informing robot design to set realistic expectations through the use of design metaphors. We contributed a methodology for determining the design metaphors of a given robot embodiment, set up a database of 165 socially interactive robot embodiments, and collected a rich dataset of participant responses about social, functional, and other relevant expectations for of those embodiments.  The proposed set of features can inform novel robot designs and compare them with extant designs through visualization techniques. The analysis results offer general guidelines for designing socially interactive robots for different contexts and ways in which user expectation of functional and social capabilities are impacted by robot embodiments.

\newpage
\bibliographystyle{ACM-Reference-Format}
\bibliography{main.bib}

\newpage
\appendix

\section{Survey Information}
\subsection{Demographic Information}\label{demographicsinfo}

\begin{figure*}[h]
     \includegraphics[width=.95\textwidth]{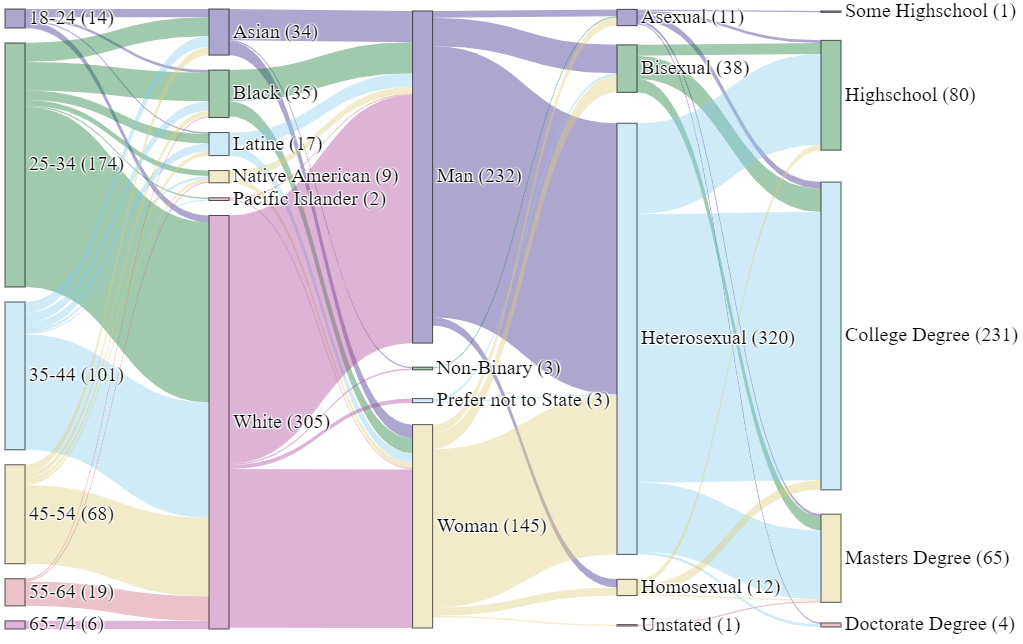}
     \label{fig:socialRole}
     \caption{The intersection of participant identities from the design metaphor survey (Study 1).}
     \label{fig:Demographics}
\end{figure*}

 \begin{figure*}[h]
     \centering
     \includegraphics[width=.85\textwidth]{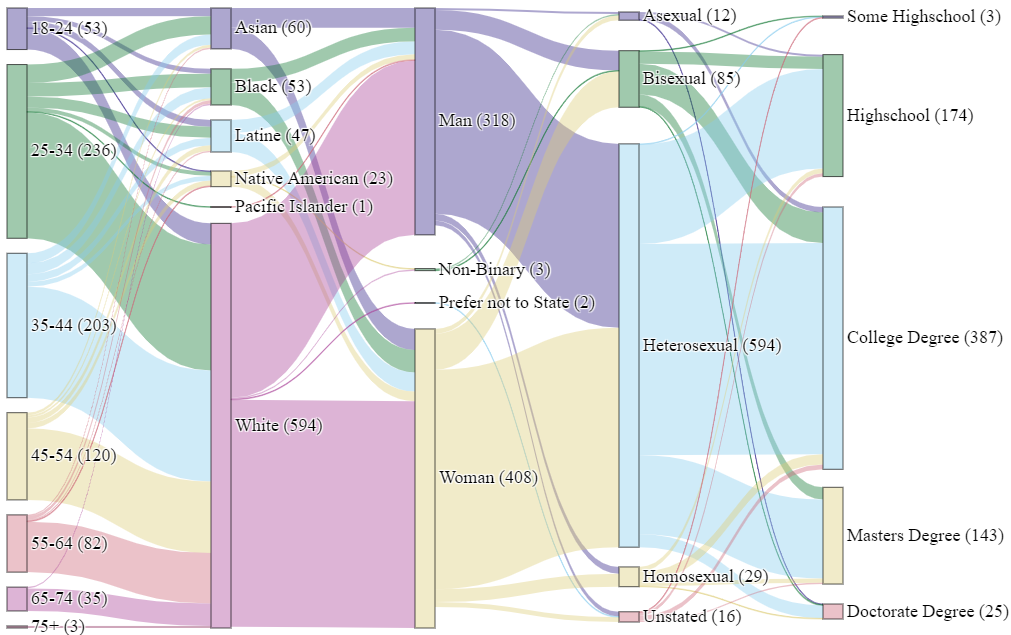}
     \caption{The intersection of participant identities from the social expectation survey (Study 2).}
     \label{fig:TI}
 \end{figure*}
     
 \begin{figure*}[h]
     \centering
     \includegraphics[width=.95\textwidth]{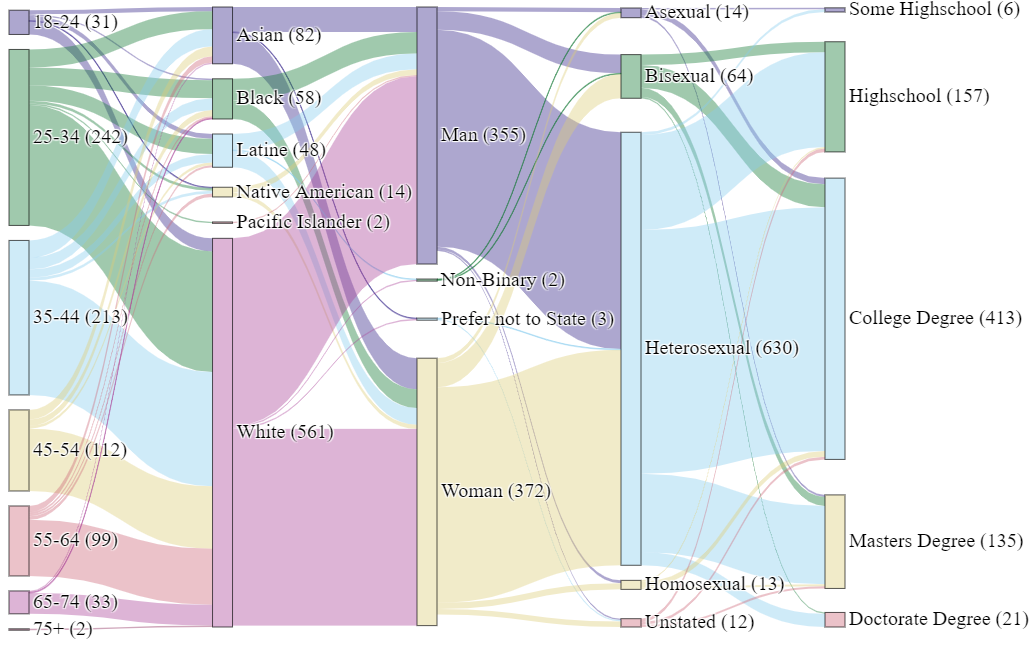}
     \caption{The intersection of participant identities from the functional expectation survey (Study 3).}
     \label{fig:socialRole}
 \end{figure*}

\newpage
\hfill\\
\newpage
\subsection{Design Metaphor Survey Questions and Interface}\label{DesignSurveyQuestions}

\begin{figure*}[h]
     \includegraphics[width=.95\textwidth]{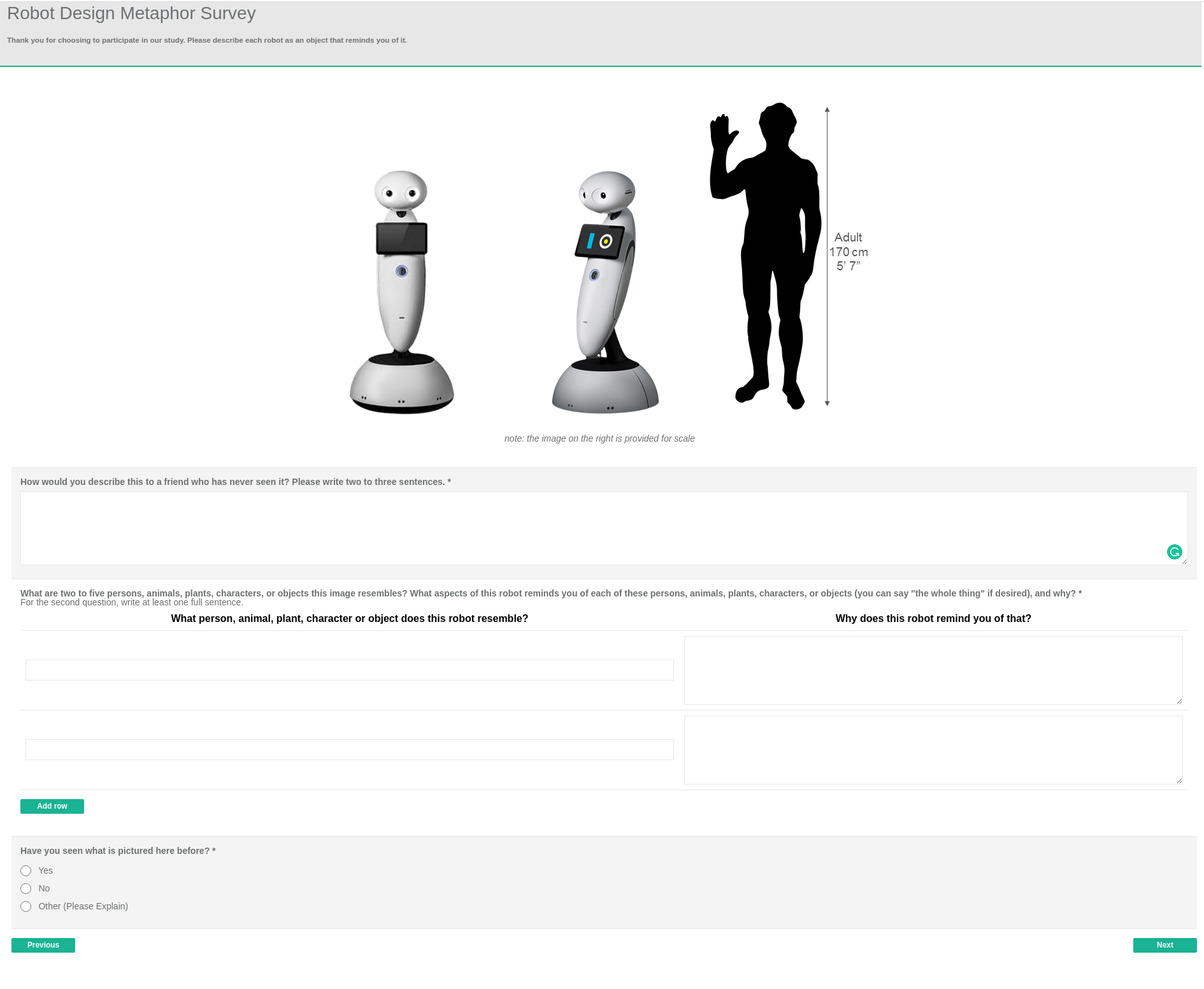}
     \label{fig:socialRole}
     \caption{The interface and questions participants in the design metaphor survey saw (Study 1).}
     \label{fig:Demographics}
\end{figure*}

\hfill\\
\newpage
\subsection{Social Expectation Survey Questions and Interface}\label{SocialSurveyQuestions}
 \begin{figure*}[h]
     \centering
     \includegraphics[width=.85\textwidth]{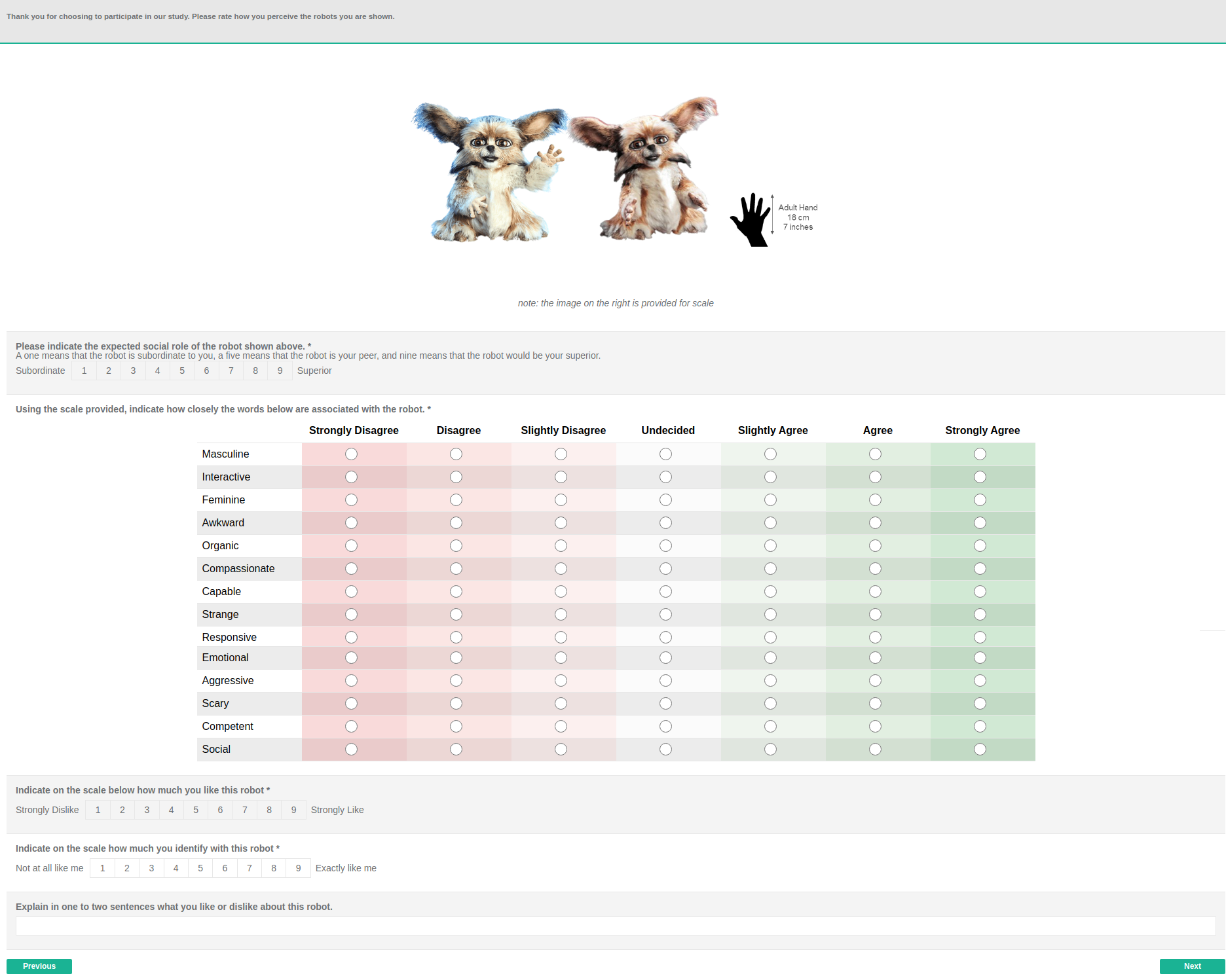}
     \caption{The interface and questions participants in the social expectation survey saw (Study 2). The order of the questions in the Likert section were randomized both during pilot studies and during the full deployment.}
     \label{fig:TI}
 \end{figure*}

\begin{center}
\begin{longtable}{|p{4cm}|p{9.5cm}|}
\caption{This table describes the assignment of the Likert items to the specific social constructs they measured.} \label{tab:long} \\

\hline \multicolumn{1}{|c|}{\textbf{Construct}} & \multicolumn{1}{c|}{\textbf{Items}}\\ \hline 
\endfirsthead

\multicolumn{2}{c}%
{{\bfseries \tablename\ \thetable{} -- Continued from previous page}} \\
\hline \multicolumn{1}{|c|}{\textbf{Construct}} & \multicolumn{1}{c|}{\textbf{Items}}\\ \hline 
\endhead

\hline \multicolumn{2}{|r|}{{Continued on next page}} \\ \hline
\endfoot

\hline \hline
\endlastfoot
 Warmth & "Social", "Organic", "Compassionate", and "Emotional"\\
 Competence & "Capable", "Responsive", "Interactive", and "Competent"\\
 Discomfort & "Scary", "Strange", "Awkward", and "Aggressive"\\
 Femininity & "Feminine"\\
 Masculinity & "Masculine"\\
\end{longtable}
\end{center}

\hfill\\
\newpage
\subsection{Functional Expectation Survey Questions and Interface}\label{FunctionalSurveyQuestions}
 \begin{figure*}[h]
     \centering
     \includegraphics[width=.85\textwidth]{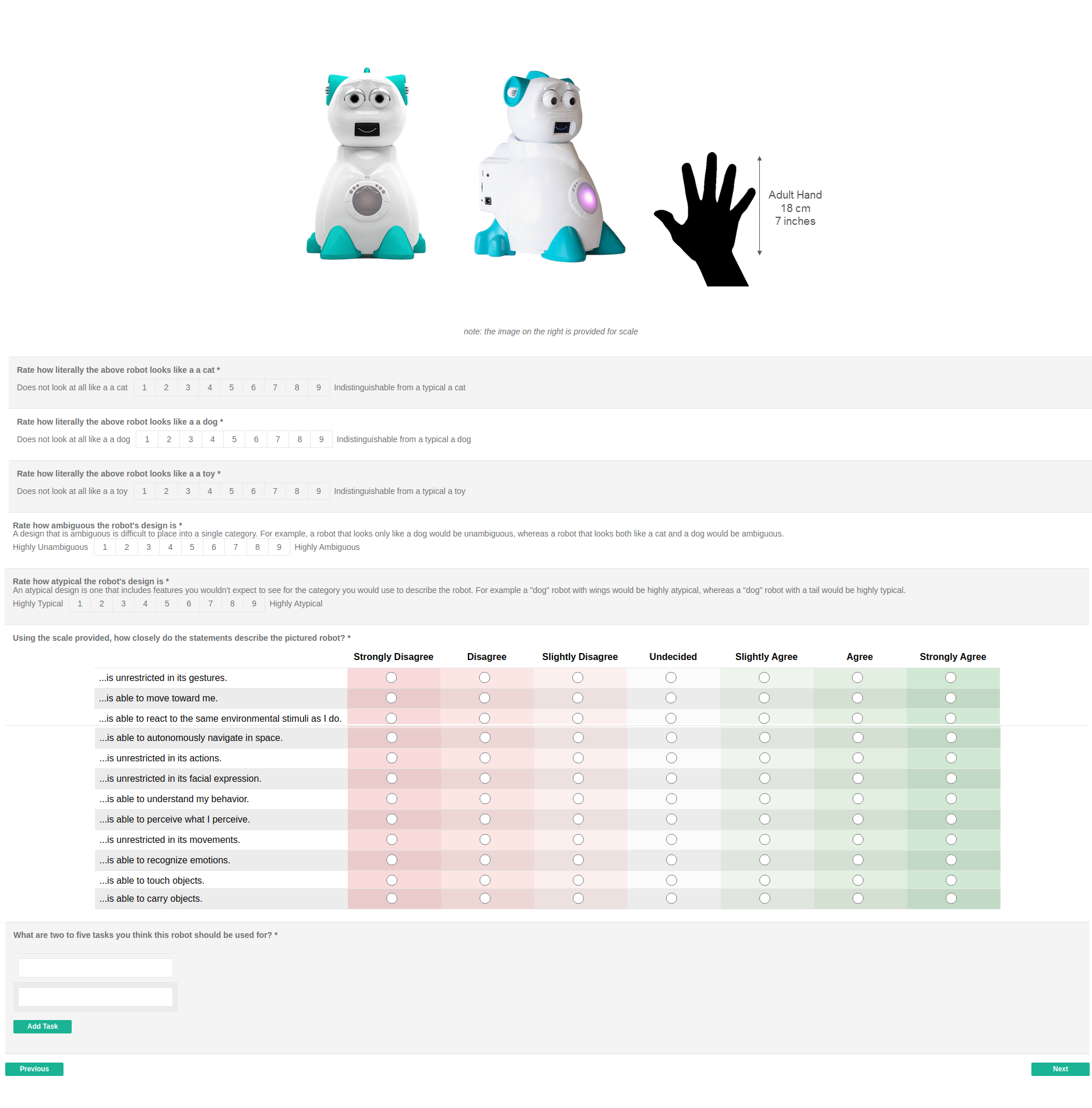}
     \caption{The interface and questions participants in the functional expectation survey saw (Study 3).}
     \label{fig:socialRole}
 \end{figure*}

\begin{center}
\begin{longtable}{|p{4cm}|p{9.5cm}|}
\caption{This table describes the assignment of the Likert items to the specific functional constructs they measured. The order of the questions in the Likert section were randomized both during pilot studies and during the full deployment.} \label{tab:long} \\

\hline \multicolumn{1}{|c|}{\textbf{Construct}} & \multicolumn{1}{c|}{\textbf{Items}}\\ \hline 
\endfirsthead

\multicolumn{2}{c}%
{{\bfseries \tablename\ \thetable{} -- Continued from previous page}} \\
\hline \multicolumn{1}{|c|}{\textbf{Construct}} & \multicolumn{1}{c|}{\textbf{Items}}\\ \hline 
\endhead

\hline \multicolumn{2}{|r|}{{Continued on next page}} \\ \hline
\endfoot

\hline \hline
\endlastfoot
 Perception and Interpretation & "...is able to react to the same environmental stimuli as I do", "...is able to recognize emotions", "...is able to perceive what I perceive", and "...is able to understand my behavior"\\
 Tactile Mobility and Interaction & "...is able to autonomously navigate in space", "...is able to move toward me", "...is able to touch objects", and "...is able to carry objects" \\
 Non-verbal Communication & "...is unrestricted in its actions", "...is unrestricted in its movements", "...is unrestricted in its facial expression", and "...is unrestricted in its gestures" \\
\end{longtable}
\end{center}

\section{Robot Descriptors}\label{RobotDescriptions}

\begin{center}
\begin{longtable}{|p{3cm}|p{8cm}|p{1.5cm}|}
\caption{A table of the binary and ordinal robot descriptors that were developed through inspection of the robot designs, and user descriptions. For each feature, we provide a description of what the feature means, and the measure of Cronbach's alpha that we obtained between two raters of the robotic systems.} \label{tab:long} \\

\hline \multicolumn{1}{|c|}{\textbf{Robot Feature}} & \multicolumn{1}{c|}{\textbf{Description}} & \multicolumn{1}{c|}{\textbf{Cronbach's $\alpha$}} \\ \hline 
\endfirsthead

\multicolumn{3}{c}%
{{\bfseries \tablename\ \thetable{} -- Continued from previous page}} \\
\hline \multicolumn{1}{|c|}{\textbf{Robot Feature}} & \multicolumn{1}{c|}{\textbf{Description}} & \multicolumn{1}{c|}{\textbf{Cronbach's $\alpha$}} \\ \hline 
\endhead

\hline \multicolumn{3}{|r|}{{Continued on next page}} \\ \hline
\endfoot

\hline \hline
\endlastfoot
 Anthropomorphic Embodiment? & Presence of human-like features (e.g., is bipedal, has two arms, two legs, or hair on the head). & .87\\
 Zoomorphic Embodiment? & Presence of animal-like features (e.g., a tail, wings, animal-like ears) & 1.00\\
 Mechanical Embodiment? & Presence of visible mechanical parts (e.g., exposed wires, wheels, or visible motors). & .89\\
 Dominant Classifaction & One of \{Anthropomorphic, Zoomorphic, Mechanical\}, which describes the overall form of embodiment. & .83\\
 
  Number of Wheels & The assumed number of wheels that the embodiment uses to move. & .70\\
  Number of Legs & The number of appendages that can be used for locomotion. & .88\\
  Number of Arms & The number of assumed appendages that could be used for gesturing and grasping. & .95\\
  Number of Eyes & The number of round components that can be perceived as eyes. & 1.00\\
  Mobile? & Can physically move between points in space. & .89 \\
  
  Does it ride on something? & Presence of a platform that the robot appears to rest on top of. & .86\\
Drivetrain Skirt? & Indicates that the wheels and motors were contained within a skirt-like shape that smoothly connects with the rest of the embodiment. & .79\\
Treads? & Presence of treads as a means of locomotion. & 1.00\\

  Spherical Head? & Presence of a head that appears to be a near-perfect sphere. & .92\\
Box Head? & Indicates that the head is approximately box-shaped (but not just a standalone screen). & .87\\
Tablet Head? & Indicates that the head consists of a single screen (e.g., a phone, tablet, etc.) & 1.00\\
Human Head? & Indicates that the head is human-like in appearance and has a skin-like quality. & 1.00\\
Wearing a Helmet? & Indicates that the robot appears to be wearing a helmet or face shield. & .61\\
Antennae? & Presence of one or more antenna-like structures on the head & 1.00\\

Hair Follicles? & Presence of many separate hair-like protrusions from the head in a distinct region that represents hair. & .87\\
Mechanical Hair? & Presence of mechanical structure on the head that can be interpreted as a hair style. & 1.00\\
Ears? & Presence of shapes or mechanisms that resemble ears. & .81\\

   Screen Face? & Presence of a screen near the top of the robot that displays at least one facial feature. & .94 \\
 Static Face? & Presence of physical facial features that are not physically actuated. & .78\\
 Mechanical Face? & Presence of a physical facial features that contains components that are physically actuated. & .77\\
 
  Mouth? & Presence of a shape or mechanism that resembles a mouth. & .89\\
  Nose? & Presence of a shape or mechanism that resembles a nose. & .83\\
  Eyebrows? & Presence of shapes or mechanisms that resemble eyebrows. & 1.00 \\
  Blush? & Presence of a shape, mechanism, or coloring that resembles rosy cheeks. & .72\\
  Eyelids? & Presence of a shape or mechanism that resembles eyelids & .72\\
Pupils? & Presence of a shape within a round shape perceived as eyes that represents a pupil. & .92\\
Irises? & Presence of a (colorful) shape within a round shape perceived as eyes that represents an iris, which contains a pupil. & .78\\
Eyelashes? & Presence of hair-like protrusions from the eye that represent eyelashes. & .89\\
Lips? & Presence of shapes or mechanisms that resemble lips. & .82\\
Mechanical Lips? & Presence of physical tube-like structures that represent lips. & 1.00 \\

 Low Waist-to-Hip Ratio? & Indicates that the perceived waist width of the robot is much smaller than ($<0.8$  times) the perceived hip width. & .80\\
High Shoulder-to-Waist Ratio? & Indicates that the perceived shoulder width is much larger than ($>1.25$ times) the perceived waist width. & .93\\
High Shoulder-Hip Ratio? & Indicates that the perceived shoulder width is much larger than ($>1.25$ times) the perceived hip width. & .62\\ 
 
 Screen On Chest? & Presence of a display interface at a medium height on the embodiment. & 1.00\\
 
Furry? & Indicates that the robot's embodiment is covered in multiple hair-like protrusions. & 1.00\\
 Matte Body? & Indicates that the external sheen of the embodiment is not highly reflective. & .94\\
Hard Exterior? & Indicates that the robot's exterior is constructed from hard materials (e.g., plastic, metal, etc.). & 1.00\\
Skin-like Material? & Indicates the presence of a skin-like, flexible, and non-furry material covering any part of the embodiment. & 1.00\\
Exposed Wires? & Presence of visible string-like structures that are needed for power requirements of the embodiment. & .80\\
Jointed Limbs? & Indicates that the limbs of the robot contain visible joints (i.e., not hidden under fabrics or outer casings). & .79\\

Industry? & Indicates that the robot was released for purchase by end-users. & .95\\

Curvy Embodiment? & Indicates that the embodiment is designed with organic-looking curves and the embodiment is not obviously partitioned into simple shapes (e.g., rectangular prisms or cylinders). & .73\\
Symmetric Embodiment? & Indicates that the embodiment exhibits reflective symmetry across its sagittal plane. & .79\\

\end{longtable}
\end{center}

\begin{center}
\begin{longtable}{|p{4cm}|p{9.5cm}|}
\caption{A table of the continuous feature descriptors taken from the robots' websites.} \label{tab:long} \\

\hline \multicolumn{1}{|c|}{\textbf{Robot Feature}} & \multicolumn{1}{c|}{\textbf{Description}}\\ \hline 
\endfirsthead

\multicolumn{2}{c}%
{{\bfseries \tablename\ \thetable{} -- Continued from previous page}} \\
\hline \multicolumn{1}{|c|}{\textbf{Robot Feature}} & \multicolumn{1}{c|}{\textbf{Description}} \\ \hline 
\endhead

\hline \multicolumn{2}{|r|}{{Continued on next page}} \\ \hline
\endfoot

\hline \hline
\endlastfoot
 Height & The total height of the robot in centimeters.\\
 Weight & The total mass of the robot in kilograms, or "UNK" if this information was not available.\\
Year & The year in which the robot was created or first written about publicly.\\
 Country of Origin & The country in which the robot was developed\\
 Most Prominent Color & The color that is used in most of the embodiment.\\
\end{longtable}
\end{center}
\end{document}